\documentclass{article}

\PassOptionsToPackage{numbers, compress}{natbib}

\usepackage[final]{neurips_fitml_2024}

\usepackage{amsmath}
\usepackage{wrapfig}
\usepackage[utf8]{inputenc} 
\usepackage[T1]{fontenc}    
\usepackage{url}            
\usepackage{booktabs}       
\usepackage{amsfonts}       
\usepackage{nicefrac}       
\usepackage{microtype}      
\usepackage{xcolor}         
\usepackage{graphicx} 
\usepackage{float}
\usepackage{multirow} 

\let\proof\relax
\let\endproof\relax

\usepackage{amsthm}
\newtheorem{theorem}{Theorem}

\newtheorem{fact}{Fact}

\newtheorem{remark}{Remark}[theorem]

\title{Generalizing Alignment Paradigm of \\ Text-to-Image Generation  with Preferences \\ through $f$-divergence Minimization}

\author{%
  Haoyuan Sun, Bo Xia, Yongzhe Chang$^*$, Xueqian Wang\thanks{Corresponding Authors
    } \\Tsinghua Shenzhen International Graduate School, Tsinghua University\\
    \{sun-hy23, xiab21\}@mails.tsinghua.edu.cn; \{changyongzhe, wang.xq\}@sz.tsinghua.edu.cn
}

\begin{document}

\maketitle

\begin{abstract}
 Direct Preference Optimization (DPO) has recently expanded its successful application from aligning large language models (LLMs) to aligning text-to-image models with human preferences, which has generated considerable interest within the community. However, we have observed that these approaches rely solely on minimizing the reverse Kullback-Leibler divergence during alignment process between the fine-tuned model and the reference model, neglecting the incorporation of other divergence constraints. In this study, we focus on extending reverse Kullback-Leibler divergence in the alignment paradigm of text-to-image models to $f$-divergence, which aims to garner better alignment performance as well as good generation diversity. We provide the generalized formula of the alignment paradigm under the $f$-divergence condition and thoroughly analyze the impact of different divergence constraints on alignment process from the perspective of gradient fields. We conduct comprehensive evaluation on image-text alignment performance, human value alignment performance and generation diversity performance under different divergence constraints, and the results indicate that alignment based on \textit{Jensen-Shannon divergence} achieves the best trade-off among them. The option of divergence employed for aligning text-to-image models significantly impacts the trade-off between alignment performance (especially human value alignment) and generation diversity, which highlights the necessity of selecting an appropriate divergence for practical applications.
\end{abstract}

\section{Introduction}
Text-to-image generative models have witnessed significant advancements in recent years \citep{podellsdxl,esser2024scaling,li2024autoregressive,betker2023improving}. When presented with appropriate textual prompts, they are capable of generating high-fidelity images that are semantically coherent with the provided descriptions, which spans a diverse range of topics, piquing significant public interest in their potential applications and societal implications. Existing self-supervised pre-trained generators, although advanced, still exhibit imperfections, with a significant challenge being their alignment with human preferences \cite{liu2021self}. 

Reinforcement Learning from Human Feedback (RLHF) has established itself as a pivotal research endeavor, demonstrating notable efficacy in aligning text-to-image models with human preferences \cite{kirstain2023pick,xu2024imagereward,blacktraining}. Faced with the intricate challenge of defining an objective that authentically encapsulates human preferences in the realm of Reinforcement Learning from Human Feedback (RLHF), researchers conventionally assemble a dataset to mirror such preferences through comparative assessments of model-generated outputs \cite{kirstain2023pick,wu2023human}. Then, a reward model is trained based on Bradley-Terry model \cite{19ff28b9-64f9-3656-ba40-08326a05748e}, inferring human preferences from the collected dataset. And the text-to-image model is fine-tuned with a reinforcement learning (RL) pipeline. It is noteworthy that such process is conducted while ensuring the model remains closely with its original form, which is achieved by employing a \textit{reverse Kullback-Leibler} divergence penalty. Significant complexity has been introduced to the RLHF pipeline due to the requirement to train a separate reward model, even though it is somewhat effective. Moreover, Reinforcement learning pipelines also present notable challenges in terms of stability and memory demands towards alignment process of text-to-image models.

Recent research has demonstrated significant success in fine-tuning large language models (LLMs) using methods based on implicit rewards, specially the Direct Preference Optimization (DPO) \cite{rafailov2023direct}. Application of similar fine-tuning techniques to text-to-image models has also produced promising results, such as Diffusion-DPO \cite{wallace2024diffusion}, D3PO \cite{yang2024using}. Such results have raisen significant interest within the community regarding the alignment of text-to-image models with human value through the methodology of utilizing implicit rewards. Furthermore, researchers have devoted significant efforts to applying such paradigm of aligning human value to text-to-image models, including SPO \cite{liang2024step}, NCPPO \cite{gambashidze2024aligning}, DNO\cite{tang2024tuning}, and so on. However, it is the situation that existing research of text-to-image generation alignment predominantly targets solutions subject to the constraint of the \textit{reverse Kullback-Leibler} divergence, with notable underexploitation of strategies that integrate other types of divergences. 

It has been pointed out that models would overfit due to repeated fine-tuning on a few images, thus leading to reduced output diversity \cite{Ruiz_2023_CVPR}. In the alignment of large language models, similar challenges exist; and some studies \cite{wiher2022decoding,perez2022red} have highlighted that the mode-seeking property of reverse KL divergence tends to reduce diversity in generated outputs, which can constrain the model's potential. Studies on aligning large language models \cite{go2023aligning,wangbeyond} indicate that the problem of diversity reduction caused by fine-tuning can be alleviated by incorporating diverse divergence constraints. Therefore, in this study, we also explore the effects of employing diverse divergence constraints on the generation diversity. 

\begin{wrapfigure}{l}{0.5\textwidth}
\centering
\includegraphics[width=0.5\textwidth]{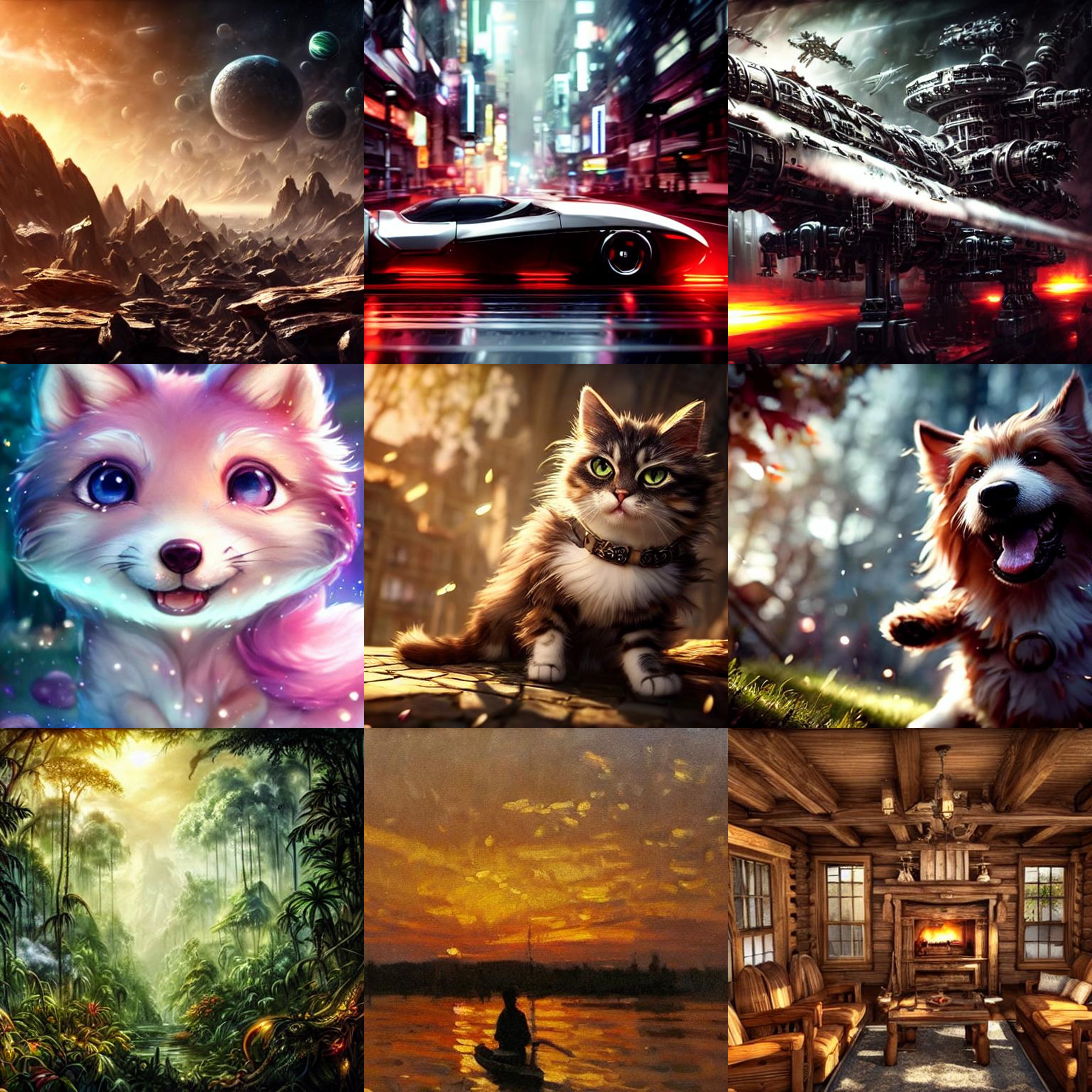}
\caption{Examples of image generated by the model aligned using the Jensen-Shannon divergence constraint.}
\label{examples}
\vspace{-1em}
\end{wrapfigure}

In this study, we generalize the alignment of text-to-image models based on \textit{reverse Kullback-Leibler} divergence to a framework based on $\mathit{f}$\textit{-divergence} constraints, which encompasses a wider range of divergences, including Jensen-Shannon divergence, forward Kullback-Leibler divergence, $\alpha$-divergence, and so on. We comprehensively
analyze the impact of diverse divergence constraints on the alignment process from the perspective of gradient fields. Furthermore, we set Step-aware Preference Optimization (SPO) \cite{liang2024step} as our benchmark method, utilize Stable Diffusion V1.5 \cite{Rombach_2022_CVPR} as our benchmark model, and assess on the test split of HPS-V2 \cite{wu2023human} with different divergence constraints. Evaluations are carried out to examine the performance of image-text alignment, human value alignment, and generation diversity, which also aim to discern the certain divergence most effectively balances these three aspects. Our results indicate that \textit{Jensen-Shannon divergence} successfully strikes the ideal equilibrium among the three criteria examined, while also achieving the highest standard in human value alignment performance. Therefore, in text-to-image alignment, judicious selection of the divergence constraint, tailored to the specific alignment requirements, is paramount. In Figure \ref{examples}, we present several images generated by the model that have been aligned under the Jensen-Shannon divergence.

To the best of our knowledge, this is the first work to apply different divergence constraints to text-to-image alignment paradigm. Our contributions are summarized as follows: (1) \textit{Generalized alignment formula}: we propose a generalized formula for text-to-image generation alignment, aiming to provide more choices on divergence constraints in alignment execution. (2) \textit{Thorough alignment process analysis}: we comprehensively analyze the impact of different divergence constraints on alignment process from the perspective of gradient fields. (3) \textit{Extensive alignment evaluations:} we conducted extensive evaluations on text-to-image generation alignment, meticulously assessing both alignment performance (image-text alignment and human value alignment) and generation diversity.

\section{Related Work}
\subsection{Aligning Text-to-Image Model with Preferences} 
Recently, inspired by the alignment approaches based on human preferences, notably exemplified by methods such as direct preference optimization (DPO) \cite{rafailov2023direct}, eliminating the need for explicit reward models and showing their significant success on Large Language Models (LLMs), and then garnering substantial attention within the community on the development of offline alignment for text-to-image diffusion models. Diffusion-DPO \cite{wallace2024diffusion} enables text-to-image diffusion models to directly learn from human feedback in an open-vocabulary setting, and fine-tunes them on the contains Pick-a-Pic \cite{kirstain2023pick} dataset with image preference pairs. Direct Preference for Denoising Diffusion Policy Optimization (D3PO) \cite{yang2024using} proposes a method on generating pairs of images from the same prompt and identifying the preferred and dispreferred images with the help of human evaluators. Step-aware Preference Optimization (SPO) \cite{liang2024step} propose an approach that preferences at each step should be assessed and it utilizes a step-aware preference model and a step-wise resampler to ensure accurate step-aware preference alignment. DenseReward method \cite{yangdense} proposes enhancing the DPO scheme by incorporating a temporal discounting approach, which prioritizes the initial denoising steps. Noise-Conditioned Perceptual Preference Optimization (NCPPO) \cite{gambashidze2024aligning} proposes that the optimization process should aligns with human perceptual features, instead of the less informative pixel space. Direct Noise Optimization (DNO) \cite{tang2024tuning} optimizes noise during the sampling process of text-to-image diffusion models. PopAlign \cite{li2024popalign} is an approach for population-level preference optimization, mitigating the biases of pretrained text-to-image diffusion models. Diffusion-KTO \cite{li2024aligning} generalizes the human utility maximization framework to the alignment of text-to-image diffusion models. While these studies have demonstrated impressive results in addressing the text-to-image alignment challenge, we also notice that they all rely on \textit{reverse Kullback-Leibler divergence} to minimize the discrepancy between the fine-tuned model and the reference model.
\subsection{$f$-divergence utilized in Generation Models}
In previous studies, researchers have extensively examined the application of $f$-divergences in generative models. In the classical work done by \cite{goodfellow2014generative}, the concept of Generative Adversarial Networks (GANs) and their relationship to the Jensen-Shannon divergence are introduced. $f$-GAN \cite{NIPS2016_cedebb6e} proposes that the variational expression of the $f$-divergence can be regarded as the loss function for Generative Adversarial Networks (GANs). Wasserstein-GAN \cite{arjovsky2017wasserstein} offers theoretical insights into the connection between the choice of divergences and the convergence of probability distributions. Moreover, in the work \cite{theis2016note}, it is proposed that utilizing various divergences and metrics can result in divergent trade-offs, and distinct evaluations tend to favor specific models. The application of $f$-divergence has also been observed in large language model alignment tasks. $f$-DPG \cite{go2023aligning} shows that Jensen-Shannon divergence strikes a good balance between different competing objectives, and often significantly outperforming the reverse Kullback-Leibler divergence. $f$-DPO \cite{wangbeyond} generalizes the framework of DPO by incorporating diverse divergence constraints; and it shows that by adjusting the divergence regularization, we can achieve a better balance between the alignment performance and the generation diversity.

\section{Preliminary}
\label{Preliminary}
\subsection{$f$-divergence}
For any convex function $f(x):\mathbb{R}^{+} \rightarrow \mathbb{R}$ with $f(1)=0$, and $p_1,p_2$ are two distributions over a discrete set $\mathcal{X}$, the $f$-divergence between $p_1$ and $p_2$ can be defined as \cite{liese2006divergences}:
\[
D_f(p_1||p_2)=\mathbb{E}_{x\sim p_2}\left[ f\left( \frac{p_1(x)}{p_2(x)}\right)+f'(\infty)p_1(p_2=0)  \right],
\]
where $f'(\infty)= \lim \limits_{t\rightarrow 0} t f(\frac{1}{t})$ \cite{hiriart1996convex}, $p_1(p_2=0)=0$ is the $p_1$-mass of the set \{$x\in\mathcal{X}:p_2(x)=0$\}. Under normal circumstances, we can make the assumption that the support set of $p_1$ is dominated by the support set of $p_2$, i.e. $Supp(p_1)\subset Supp(p_2)$, and then we can have  $p_1(p_2=0)=0$. Hence, the aforementioned definition can be simplified as:
\[
    D_f(p_1||p_2)=\mathbb{E}_{x\sim p_2}\left[ f\left( \frac{p_1(x)}{p_2(x)}\right)\right]
\]
For different functions $f(x)$, the $f$-divergence class encompasses a wide range of commonly employed divergence measures, such as reverse Kullback-Leibler (KL) divergence, forward Kullback-Leibler (KL) divergence, $\alpha$-divergence ($\alpha \in (0,1)$), Jensen-Shannon (JS) divergence, and so on.
\begin{wraptable}{l}{0.7\textwidth} 
\vspace{-0.5em}
    \centering
    \begin{tabular}{c|ccc}
\toprule
\small{$f-$divergence} & $f(x)$ & $f'(x)$ & $f''(x)$ \\ \midrule
\small{Reverse KL}   & \small{$ x\log x$} & \small{$\log x + 1$} & \small{$\frac{1}{x}$} \\ \midrule
\small{Forward KL}   & \small {$-\log x$} & $-\frac{1}{x}$ & \small{$\frac{1}{x^{2}}$} \\ \midrule
\small{$\alpha$-divergence} & $\frac{x^{1-\alpha}-(1-\alpha)x-\alpha}{\alpha(\alpha-1)}$  & $\frac{1-x^{-\alpha}}{\alpha}$ & \small{$\frac{1}{x^{\alpha+1}}$} \\ \midrule
\small{JS divergence} & \small{ $x \log \frac{2x}{x+1}+\log\frac{2}{x+1}$ }& \small{$\log \frac{2x}{1+x}$} & \small{$\frac{1}{x(1+x)}$}\\
 \bottomrule
\end{tabular}
    \caption{Several commonly used $f$-divergence with their derivatives and second derivatives. }
    \label{f-divergence}
    \vspace{-0.5em}
\end{wraptable}
In previous studies, reverse KL divergence can be regarded as a specific instance of $\alpha$-divergence with $\alpha=0$; and forward KL divergence as a specific instance of $\alpha$-divergence with $\alpha=1$. We summarize several commonly used $f$-divergence, the derivatives and the second derivatives in Table \ref{f-divergence}.
\\

\section{Method}
Much like in the alignment tasks of large language models, there are many concepts that are analogous in the alignment tasks of text-to-image models, and we start by elucidating these parallels. Firstly, the question input of LLMs is akin to the text (condition) input of T2I models, i.e. $x \rightarrow c$; and the output answer of LLMs is akin to the generated image of T2I models, i.e. $y \rightarrow x_0$. Moreover, the policy of LLMs parallels the sampling probability of T2I models (especially diffusion models), i.e. $\pi(y|x) \rightarrow p(x_{0:T}|c)$. Finally, the preference data for output answers of LLMs is analogous to the preference data for generated images of T2I models, i.e.  $(x,y_w,y_l)\rightarrow (c,x_0^w,x_0^l)$. In the following subsections, we first derive the generalized formula of alignment objective function. Then, we analyze the gradient field of different divergences on the alignment process with respect to the objective function and comprehensively analyze the impact of diverse divergence constraints on alignment performance. 
\subsection{Generalized Formula}
In previous studies of Reinforcement Learning from Human Feedback (RLHF), researchers typically aim to maximize the reward function ($r(c,x_{0:T})$) while penalizing the reverse KL divergence between the fine-tuned model and the original model to prevent it from collapsing during training. In our study, we generalize such penalty constraint from the reverse KL divergence ($D_{\text{KL}}\left(p_{\theta}(x_{0:T}|c),p_{\text{ref}}(x_{0:T}|c)\right)$) to the $f$-divergence ($D_{f}\left(p_{\theta}(x_{0:T}|c),p_{\text{ref}}(x_{0:T}|c)\right)$).

We reframe the reinforcement learning objective function as an optimal problem, presenting its formulation as follows:
\[
\begin{split}
\arg\max\limits_{p_{\theta}} \mathbb{E}_{c\sim p_c,x_{0:T}\sim p_{\theta}(x_{0:T}|c)}\big[r(c,x_{0:T})\big]
&-\beta D_f\Big(p_{\theta}(x_{0:T}|c),p_{\text{ref}}(x_{0:T}|c)\Big) \\
\text{s.t.}\quad \sum \limits_{x_{0:T}}p_{\theta}(x_{0:T}|c)=1  &; \forall x_{0:T} \quad
p_{\theta}(x_{0:T}|c)\geq 0
\end{split}
\]
Such optimization problem can be addressed through the Karush-Kuhn-Tucker (KKT) conditions. Firstly, according to the definition of $f$-divergence, we construct the following Lagrangian function: 
\[
\begin{split}
\mathcal{L}(p_{\theta}(x_{0:T}|c),\lambda,\zeta(x_{0:T}))
=\mathbb{E}_{c\sim p_c,x_{0:T}\sim p_{\theta}}\big[&r(c,x_{0:T})\big]
-\beta \mathbb{E}_{p_{\text{ref}}} f\Bigg( \frac{p_{\theta}(x_{0:T}|c)}{p_{\text{ref}}(x_{0:T}|c)} \Bigg)  \\ &- \lambda \left(\sum \limits_{x_{0:T}}p_{\theta}(x_{0:T})-1\right)
+\sum \limits_{x_{0:T}} \zeta(x_{0:T})p_{\theta}(x_{0:T}|c)
\end{split}
\]
Furthermore, we can derive the Theorem \ref{f-Diffusion-DPO} from the \textit{Stationarity Condition} and \textit{Complementary Slackness} of the Karush-Kuhn-Tucker (KKT) conditions, i.e.
\[
\left\{
\begin{aligned}
&\nabla_{p_{\theta}(x_{0:T}|c)}\mathcal{L}(p_{\theta}(x_{0:T}|c),\lambda,\zeta(x_{0:T}))=0 ;\\
&\forall x_{0:T}, \zeta(x_{0:T})p_{\theta}(x_{0:T}|c) = 0.
\end{aligned}
\right.
\]
\begin{theorem}
\label{f-Diffusion-DPO}
If $p_{\mathrm{ref}}(x_{0:T}|c) > 0$ holds for all condition $c$, $f'(x)$ is an invertible function and $0$ is not in the definition domain of function $f'(x)$, the reward class consistent with Bradley-Terrry model can be reparameterized with the sampling probability $p_{\theta}(x_{0:T}|c)$ and the reference sampling probability $p_{\mathrm{ref}}(x_{0:T}|c)$ as:
\begin{align}
\label{r(c,x_0)}
    r(c,x_{0:T})=\beta f'\Bigg(\frac{p_{\theta}(x_{0:T}|c)}{p_{\text{ref}}(x_{0:T}|c)}\Bigg) + \mathrm{const}
\end{align}
\end{theorem}

As shown in Theorem \ref{f-Diffusion-DPO}, the reward function can be represented by a sampling probability $p_{\theta}(x_{0:T}|c)$, a reference sampling probability $p_{\text{ref}}(x_{0:T}|c)$, and a constant $\lambda$ that is independent of $x_{0:T}$.
Finally, substituting Equation \eqref{r(c,x_0)} into the Bradley-Terrry model \cite{19ff28b9-64f9-3656-ba40-08326a05748e} enables us to derive the generalized formula of text-to-image generation with preferences in Theorem \ref{f-Diffusion-DPO-corollary}.
\begin{theorem}
\label{f-Diffusion-DPO-corollary}
In the substitution process of Bradley-Terry model, the constant
 $ \lambda $ is independent of
$x_{0:T}$ and thus can be canceled out, resulting in the following form:
\begin{equation}
\mathcal{L}(\theta)=\mathbb{E}_{\substack{(c,x_0^{w},x_0^{l})\sim \mathcal{D},\\x_{1:T}^{w}\sim p_{\theta}(x_{1:T}^{w}|x_0^w,c),\\x_{1:T}^{l}\sim p_{\theta}(x_{1:T}^{l}|x_0^l,c).}} 
-\log \sigma\left[\beta f' \left(\frac{p_{\theta}(x_{0:T}^{w}|c)}{p_{\text{ref}}(x_{0:T}^{w}|c)}\right)-\beta f' \left( \frac{p_{\theta}(x_{0:T}^{l}|c)}{p_{\text{ref}}(x_{0:T}^{l}|c)} \right)\right]
\end{equation}
where $\sigma(\cdot)$ is the Sigmoid function; $f'(\cdot)$ represents the derivatives of $f(\cdot)$, as listed in Table \ref{f-divergence}; $\beta$ is the penalty coefficient.
\end{theorem}

So far, we have derived the generalized formula for text-to-image generation alignment with preferences. With different divergence constraint choices, we can obtain diverse alignment objectives, thereby offering more options for the alignment process.
\subsection{Analysis on Gradient Fields of Alignment Process}

In this section, we delve into the gradient fields of alignment objective functions derived from various $f$-divergence, which aims to further elucidate the intricate mechanisms underlying the alignment process. 

Let's abstract from the specific details of $f'(\cdot)$, and concentrate instead on a more general formulation of the loss function:
\begin{align}
\label{generalized loss function}
    \mathcal{L}_f\left( \mathrm{X}_1, \mathrm{X}_2 \right)=-\mathbb{E}\Big[\log \sigma \big( \beta f'(\mathrm{X}_1)-\beta f'(\mathrm{X}_2) \big)\Big]
\end{align}
where $\mathrm{X_1}$ is the training win ratio, and is equivalent to $\frac{p_{\theta}(x_{0:T}^{w}|c)}{p_{\text{ref}}(x_{0:T}^{w}|c)}$ ; similarly, $\mathrm{X}_2$ is the training loss ratio, and is identical to $\frac{p_{\theta}(x_{0:T}^{l}|c)}{p_{\text{ref}}(x_{0:T}^{l}|c)}$. We present the gradients of Equation \eqref{generalized loss function} with respect to $\mathrm{X_1}$ and $\mathrm{X_2}$ in the ensuing Theorem \ref{gradients_divergence}:
\begin{theorem}
    The partial derivatives (gradients) of $\mathrm{X_1}$ and $\mathrm{X_2}$ resulting from Equation \eqref{generalized loss function} can be expressed as follows:
\[
\left\{
\begin{aligned}
\frac{\partial \mathcal{L}_f(\mathrm{X_1},\mathrm{X_2})}{\partial \mathrm{X_1}}=-\beta \left(1-\sigma \left( 
\beta f'(\mathrm{X}_1)-\beta f'(\mathrm{X}_2) \right)\right)f''(\mathrm{X}_1)\\
\frac{\partial \mathcal{L}_f(\mathrm{X_1},\mathrm{X_2})}{\partial \mathrm{X_2}}=\beta \left(1-\sigma \left( 
\beta f'(\mathrm{X}_1)-\beta f'(\mathrm{X}_2) \right)\right)f''(\mathrm{X}_2)
\end{aligned}
\right.
\]
Thus, the gradient ratio of $\mathcal{L}_f(\mathrm{X_1},\mathrm{X_2})$ between enhancement in probability for human-preferred responses ($\mathrm{X}_1$) and reduction in probability for human-dispreferred responses ($\mathrm{X}_2$) has the expression:
\begin{align}
    \label{gradient ratio}
        \Bigg|  \frac{\partial \mathcal{L}_f(\mathrm{X_1},\mathrm{X_2})}{\partial \mathrm{X_1}} /  \frac{\partial \mathcal{L}_f(\mathrm{X_1},\mathrm{X_2})}{\partial \mathrm{X_2}} \Bigg| = \frac{f''(\mathrm{X}_1)}{f''(\mathrm{X}_2)}
\end{align}
\label{gradients_divergence}
\end{theorem}
Referencing Table \ref{f-divergence}, different divergences yield distinct gradient ratios. If selected divergence is reverse Kullback-Leibler divergence, the gradient ratio is $\frac{\mathrm{X}_2}{\mathrm{X}_1}$; if selected divergence is Jensen-Shannon divergence, the gradient ratio is $\frac{\mathrm{X}_2\cdot (\mathrm{X}_2+1)}{\mathrm{X}_1 \cdot (\mathrm{X}_1+1)}$; if selected divergence is $\alpha$-divergence, the gradient ratio is $\frac{\mathrm{X}_2^{1+\alpha}}{\mathrm{X}_1^{1+\alpha}}$; if selected divergence is forward Kullback-Leibler divergence, the gradient ratio is $\frac{\mathrm{X}_2^{2}}{\mathrm{X}_1^{2}}$. Previous studies \cite{feng2024towards,yan20243d} present the results of original DPO framework, focusing particularly on its application in the context of reverse Kullback-Leibler divergence; while our outcomes show the generalization under diverse divergences.   

Furthermore, as the alignment advances, the value of $\mathrm{X}_1$ tends to increase to more than 1, whereas $\mathrm{X}_2$ tends to decrease to less than 1. Hence, for any pairwise preference data, $\mathrm{X}_2 / \mathrm{X}_1 < 1$ holds during the alignment process. Then, Theorem \ref{divergence-compare} can be easily derived.
\begin{theorem}
\label{divergence-compare}
As alignment progresses, we have $\mathrm{X}_2 / \mathrm{X}_1 < 1$.  Hence,
    \[
    0 < \frac{\mathrm{X}_2^2}{\mathrm{X}_1^2} < \frac{\mathrm{X}_2 \cdot (\mathrm{X}_2+1)}{\mathrm{X}_1 \cdot (\mathrm{X}_1+1)}<\frac{\mathrm{X}_2}{\mathrm{X}_1} < 1 \, \; \, \, \text{and} \, \; \, \,
    0 < \frac{\mathrm{X}_2^2}{\mathrm{X}_1^2}  < \frac{\mathrm{X}_2^{1.8}}{\mathrm{X}_1^{1.8}}  < \frac{\mathrm{X}_2^{1.6}}{\mathrm{X}_1^{1.6}}  < \frac{\mathrm{X}_2^{1.4}}{\mathrm{X}_1^{1.4}}  < \frac{\mathrm{X}_2^{1.2}}{\mathrm{X}_1^{1.2}}  < \frac{\mathrm{X}_2}{\mathrm{X}_1} < 1
    \]
\end{theorem}

Theorem \ref{divergence-compare} presents the inequality of gradient ratio of different divergences. A lower gradient ratio results in a swifter alteration in the probability of a dispreferred image compared to that of a preferred one, indicating a more pronounced decrease in the probability of dispreferred images.
\noindent Hence, the decline varies in intensity, with \textit{forward KL divergence ($\alpha$=1)} exhibiting the highest decrease, \textit{reverse KL divergence ($\alpha$=0)} the lowest, and both \textit{Jensen-Shannon divergence} and \textit{$\alpha$-divergence ($\alpha \in$(0,1))} falling in between.

In order to obtain a more intuitive understanding of the impact of different divergence choices during the alignment process, we visualize the landscape of alignment objective functions with different divergences in Equation \eqref{generalized loss function} in Appendix \ref{Plot of Gradient Fields}.

\section{Experiments}

In this section, we present extensive experimental evaluations to answer the following questions:

\textbf{Q1}: When choosing different divergence constraints, would it have a significant impact on the final \textit{image-text} alignment performance?

\textbf{Q2}: When choosing different divergence constraints, would it have a significant impact on the alignment of \textit{human value}? If so, which divergence constraint achieves the best performance? 

\textbf{Q3}: When choosing different divergence constraints, would it have a significant impact on the \textit{generation diversity}? Which divergence can achieve the best trade-off between alignment performance and generation diversity?

\subsection{Experimental Settings}
\subsubsection{Benchmark.} Step-aware Preference Optimization (SPO) \cite{liang2024step} employs a step-aware preference model and a step-wise resampler to guarantee precise step-aware preference alignment. Consequently, to support a more tangible experimental assessment, we select SPO as our benchmark approach. To establish a fair basis for comparison with prior methods, we select Stable Diffusion v1-5 model \cite{Rombach_2022_CVPR} as our benchmark model. In order to conduct a more comprehensive evaluation, we utilize the test set of HPS-V2 \cite{wu2023human} as our evaluation benchmark dataset, which comprises 400 prompts. We report the mean and standard deviation of metrics of the generated image for these prompts.
\subsubsection{Evaluation Metrics.} We evaluate the generated images from three aspects (for the aforementioned three questions).

In terms of model's image-text alignment performance (for Q1), we adopt the widely used evaluation metrics in Text-to-Image models, i.e., Text-Image CLIP score \cite{hessel2021clipscore} and VQAScore \cite{lin2025evaluating}. CLIP score is fundamentally based on the CLIP model, transforming input text and images into distinct text and image vectors, and then followed by calculation of the dot product of these vectors. VQAScore, an efficacy metric, emerges from the training of generative vision-language models designed for visual-question-answering endeavors, where an image and a query converge to yield a response. Hence, higher Text-Image CLIP score and VQAScore indicate better alignment between the text and the image.

In terms of model's human value alignment performance (for Q2), we adopt four metrics for comprehensive evaluation. Aesthetic score is obtained using the LAION Aesthetics Predictor \cite{schuhmann2022laion}, which quantifies the average human appreciation for the visual appeal of generated images. ImageReward \cite{xu2024imagereward}, leveraging a structure that
combines ViT-L for image encoding and a 12-layer Transformer for text encoding, which effectively models the human value and preference. PickScore \cite{kirstain2023pick} is an advanced scoring function built upon a meticulously curated comprehensive dataset named "Pick-a-Pic". Human Preference Score v2 (HPS-v2) \cite{wu2023human} has been developed through the refinement of the CLIP model on HPD-v2, which enhances the precision of assessing human preferences for generated images. Furthermore, higher Aesthetics Score, ImageReward, Pickscore and HPS-V2 suggest better alignment with human value.

In terms of diversity of generated images from the aligned model (for Q3), we adopt eight metrics for a further comprehensive evaluation. Image-Image CLIP score \cite{hessel2021clipscore} serves as a reliable metric for assessing similarity between images. RMSE, PSNR, and SSIM are conventional metrics used to evaluate image similarity, we also utilize them to assess the diversity of generated images. Feature Similarity Index Measure (FSIM) \cite{zhang2011fsim} quantifies the similarity between images by assessing the alignment of edges, shapes, visual patterns, and surface attributes. Learned Perceptual Image Patch Similarity (LPIPS) \cite{zhang2018unreasonable} utilizes the feature representations learned by a deep neural network, which is capable of capturing details of human visual perception such as texture, color, and structure; then the computation of perceptual similarity between two images can be conducted. Furthermore, it's worth noting that these six metrics all initially describe the similarity between images; and when they are used to describe generation diversity, their properties are the opposite of their properties when describing similarity. Moreover, we opt for Image Entropy, encompassing both Entropy 1D and Entropy 2D, to evaluate the information content diversity within images themselves; they quantifies the average information per pixel, with higher entropy values indicative of a greater diversity and richness in the image’s information content.

\subsection{Image-Text Alignment (For Q1)}

For text-to-image models, the alignment performance between text prompt and generated images is a crucial evaluation metric. Therefore, we test the Text-Image CLIP score and VQAScore of all models fine-tuned under different divergence constraints to assess the alignment performance in Table \ref{evaluations of alignment performance}. The results indicate that the reverse Kullback-Leibler divergence achieves the best text-image alignment performance; while it is also worth noting that different divergences do not significantly affect the final text-image alignment performance.

\setlength{\tabcolsep}{1.5mm}
\begin{table*}[ht]\scriptsize
\centering
\begin{tabular}{cc|cc|cccc}
\toprule

\multicolumn{2}{c|}{Model}& \multicolumn{1}{c|}{CLIPScore $\uparrow$}& \multicolumn{1}{c|}{VQAScore $\uparrow$}&\multicolumn{1}{c|}{Aesthetics Score $\uparrow$} & \multicolumn{1}{c|}{ImageReward $\uparrow$} & \multicolumn{1}{c|}{PickScore $\uparrow$} &  {HPS-V2   $\uparrow$}                           \\ \midrule \midrule 
\multicolumn{2}{c|}{Original Model}  & 0.352$\pm$0.049 & 0.625$\pm$0.239  & 5.648$\pm$0.526                     & 0.173$\pm$1.011                   & 20.908$\pm$1.228                 & \multicolumn{1}{c}{26.933$\pm$1.454} \\ \midrule \midrule
\multicolumn{2}{c|}{Reverse KL Divergence}& \textbf{0.363$\pm$0.049} & \textbf{0.676$\pm$0.236} &5.812$\pm$0.514                     & 0.619$\pm$0.921                  & 21.621$\pm$1.151                 & 27.801$\pm$1.352                      \\ \midrule
\multicolumn{1}{c|}{\multirow{5.20}{*}{$\alpha$-Divergence}} & $\alpha$=0.2 & 0.360$\pm$0.048 &  0.673$\pm$0.228 &5.827$\pm$0.546                     & 0.561$\pm$0.957                  & 21.528$\pm$1.177                 & 27.848$\pm$1.391                      \\ \cmidrule{2-8} 
\multicolumn{1}{c|}{}                              & $\alpha$=0.4 & 0.361 
$\pm$0.047&  0.675$\pm$0.233  & 5.755$\pm$0.518                     & 0.622$\pm$0.911                  & 21.569$\pm$1.204                 & 27.762$\pm$1.385                      \\ \cmidrule{2-8} 
\multicolumn{1}{c|}{}                              & $\alpha$=0.6  & 0.358 
$\pm$0.047  &  0.657$\pm$0.232
 & 5.769$\pm$0.481                     & 0.491$\pm$0.943                  & 21.357$\pm$1.180                 & 27.712$\pm$1.350                      \\ \cmidrule{2-8} 
\multicolumn{1}{c|}{}                              & $\alpha$=0.8 & 0.361$\pm$0.050   & 0.662$\pm$0.236
 & 5.821$\pm$0.511                     & 0.561$\pm$0.965                  & 21.483$\pm$1.175                 & 27.675$\pm$1.379                      \\ \midrule
\multicolumn{2}{c|}{Forward KL Divergence} & 0.362$\pm$0.050  & 0.670$\pm$0.231 &5.844$\pm$0.528                     & 0.551$\pm$0.946                  & 21.552$\pm$1.170                 & 27.822$\pm$1.355                      \\ \midrule
\multicolumn{2}{c|}{Jensen-Shannon Divergence} & 0.361$\pm$0.049 & 0.665$\pm$0.231 & \textbf{5.884$\pm$0.514}            & \textbf{0.631$\pm$0.939}         & \textbf{21.635$\pm$1.149}        & \textbf{27.850$\pm$1.388}             \\ \bottomrule
\end{tabular}
\caption{Evaluations of the alignment performance, where the Text-Image CLIP score and VQAScore evaluate image-text alignment performance, and the remaining four metrics evaluate human value alignment performance.}
\label{evaluations of alignment performance}
\end{table*}

\subsection{Human Value Alignment (For Q2)}

Evaluating how well the aligned models are with human values and preferences is crucial. To comprehensively assess various divergences in aligning with human values, we compare their performances systematically on four metrics: Aesthetic score, ImageReward, PickScore, and HPS-V2 in Table \ref{evaluations of alignment performance}. The comparison between the results of the fine-tuned models and the original model indicates that the alignment process effectively enhances the model in terms of its performance in human values. Furthermore, in comparing the influence of diverse divergence constraints on human value alignment, the results reveal that different divergence would significantly affect human value alignment; remarkably, the \textit{Jensen-Shannon (JS) divergence} exhibits the best performance across all four human value alignment metrics, suggesting that it serves as a more potent constraint specifically for the scenario of human value alignment. Actually, it also aligns with our previous analysis of the gradient fields, where the Jenson-Shannon (JS) divergence shows the smoothest loss function surface and suboptimal gradient ratio, resulting in a more stable alignment process.
\begin{table*}[t]\scriptsize
\centering
\begin{tabular}{cc|cccc}
\toprule
\multicolumn{2}{c|}{Model}                               & \multicolumn{1}{c|}{Image-Image CLIP score $\downarrow$} & \multicolumn{1}{c|}{Entropy 1D $\uparrow$} & \multicolumn{1}{c|}{Entropy 2D $\uparrow$} & \multicolumn{1}{c}{LPIPS $\uparrow$} \\ \midrule \midrule
\multicolumn{2}{c|}{Original Model}                      & 0.8052 $\pm$ 0.0824                         & 3.8235 $\pm$ 0.2960                 & 7.5474 $\pm$ 0.6516                 & 0.2972 $\pm$ 0.0419            \\ \midrule \midrule
\multicolumn{2}{c|}{Reverse KL Divergence}               & 0.8448 $\pm$ 0.0774                         & 3.9613 $\pm$ 0.1467                 & 7.8347 $\pm$ 0.3694                 & 0.2907 $\pm$ 0.0363            \\ \midrule
\multicolumn{1}{c|}{\multirow{5.20}{*}{$\alpha$-Divergence}} & $\alpha=$0.2 & 0.8436 $\pm$ 0.0854                         & 3.9411 $\pm$ 0.1885                 & 7.7836 $\pm$ 0.4400                 & 0.3047 $\pm$ 0.0377            \\ \cmidrule{2-6} 
\multicolumn{1}{c|}{}                              & $\alpha=$0.4 & 0.8377 $\pm$ 0.0824                         & \textbf{3.9784 $\pm$ 0.1464}        & 7.8206 $\pm$ 0.3729                 & 0.2959 $\pm$ 0.0349            \\ \cmidrule{2-6} 
\multicolumn{1}{c|}{}                              & $\alpha=$0.6 & \textbf{0.8372 $\pm$ 0.0825}                & 3.9275 $\pm$ 0.1991                 & 7.7937 $\pm$ 0.4625                 & \textbf{0.3109 $\pm$ 0.0373}   \\ \cmidrule{2-6} 
\multicolumn{1}{c|}{}                              & $\alpha=$0.8 & 0.8423 $\pm$ 0.0795                         & 3.9594 $\pm$ 0.1666                 & 7.7563 $\pm$ 0.4179                 & 0.3001 $\pm$ 0.0377            \\ \midrule
\multicolumn{2}{c|}{Forward KL Divergence}               & 0.8454 $\pm$ 0.0821                         & 3.9477 $\pm$ 0.1555                 & 7.7750 $\pm$ 0.3619                 & 0.2962 $\pm$ 0.0347            \\ \midrule
\multicolumn{2}{c|}{Jensen-Shannon Divergence}           & 0.8448 $\pm$ 0.0798                         & 3.9632 $\pm$ 0.1487                 & \textbf{7.8767 $\pm$ 0.3801}        & 0.2989 $\pm$ 0.0361            \\ \bottomrule
\end{tabular}
\\
\begin{tabular}{cc|cccc}
\toprule
\multicolumn{2}{c|}{Model}                               & \multicolumn{1}{c|}{\;\quad\quad RMSE $\uparrow$\quad\quad\;} & \multicolumn{1}{c|}{\quad\quad\quad PSNR $\downarrow$\quad\quad\quad} & \multicolumn{1}{c|}{\quad\quad\quad SSIM $\downarrow$\quad\quad\quad} & \multicolumn{1}{c}{\quad\quad FSIM $\downarrow$\quad\quad} \\ \midrule \midrule
\multicolumn{2}{c|}{Original Model}                      & 0.0132 $\pm$ 0.0028           & 37.745 $\pm$ 1.843            & 0.8839 $\pm$ 0.0382           & 0.3791 $\pm$ 0.0230           \\ \midrule \midrule
\multicolumn{2}{c|}{Reverse KL Divergence}               & 0.0132 $\pm$ 0.0028           & 36.398 $\pm$ 1.573            & 0.8512 $\pm$ 0.0372           & 0.3813 $\pm$ 0.0182           \\ \midrule
\multicolumn{1}{c|}{\multirow{4}{*}{$\alpha$-Divergence}} & $\alpha=$0.2 & 0.0163 $\pm$ 0.0027           & 35.856 $\pm$ 1.467            & 0.8404 $\pm$ 0.0368           & \textbf{0.3759 $\pm$ 0.0212}  \\ \cmidrule{2-6} 
\multicolumn{1}{c|}{}                              & $\alpha=$0.4 & 0.0154 $\pm$ 0.0025           & 36.363 $\pm$ 1.427            & 0.8530 $\pm$ 0.0348           & 0.3821 $\pm$ 0.0180           \\ \cmidrule{2-6} 
\multicolumn{1}{c|}{}                              & $\alpha=$0.6 & \textbf{0.0166 $\pm$ 0.0026}  & \textbf{35.705 $\pm$ 1.373}   & \textbf{0.8357 $\pm$ 0.0349}  & 0.3778 $\pm$ 0.0209        \\ \cmidrule{2-6} 
\multicolumn{1}{c|}{}                              & $\alpha=$0.8 & 0.0155 $\pm$ 0.0028           & 36.320 $\pm$ 1.566            & 0.8517 $\pm$ 0.0374           & 0.3806 $\pm$ 0.0215           \\ \midrule
\multicolumn{2}{c|}{Forward KL Divergence}               & 0.0157 $\pm$ 0.0026           & 36.171 $\pm$ 1.457            & 0.8468 $\pm$ 0.0351           & 0.3780 $\pm$ 0.0185           \\ \midrule
\multicolumn{2}{c|}{Jensen-Shannon Divergence}           & 0.0158 $\pm$ 0.0026           & 36.104 $\pm$ 1.431            & 0.8449 $\pm$ 0.0354           & 0.3817 $\pm$ 0.0195           \\ \bottomrule
\end{tabular}
\caption{Evaluations of the generation diversity. The metrics originally utilized for evaluating image similarity exhibit an opposite property when evaluating generation diversity.}
\label{generation diversity evaluation}
\end{table*}
\subsection{Generation Diversity (For Q3)}

We evaluate the generation diversity of aligned models using different divergence constraints from multiple perspectives (embedding diversity, pixel-level diversity, structural diversity, perceptual diversity, information complexity, and so on), and the corresponding results are shown in Table \ref{generation diversity evaluation}. From the results, we can observe that different divergence constraints exhibit advantages in different aspects when evaluated with different generation diversity metrics. Firstly, we would like to compare the aligned models under different divergence constraints to the original model: it is demonstrated that the aligned models show a decrease in embedding diversity; however, they exhibit improvements in other aspects such as pixel-level diversity, structural diversity, information complexity. Such observation reveals a transformation in the alignment process where the variety of the primary subject diminishes, yet the intricacy and breadth of details and structures of the generated images expand, echoing findings from DreamBooth \cite{Ruiz_2023_CVPR}. 

Furthermore, it has also indicated that increased generative diversity is associated with a decline in alignment performance (both image-text alignment and human value alignment). Therefore, careful consideration of the trade-off between alignment performance and generation diversity is essential when choosing the divergence constraint. Through a deeper comparison and analysis, we can observe that Jensen-Shannon divergence outperforms or matches reverse Kullback-Leibler divergence across most diversity metrics. Combining such observation with the previous evaluation of alignment performance where it achieves the best human value alignment, we believe Jensen-Shannon divergence is a better trade-off between alignment performance and generation diversity.

\section{Conclusion}
In this paper, we extend the alignment framework for text-to-image models, transitioning from a criterion based on the reverse Kullback-Leibler (KL) divergence to a more inclusive framework grounded in $f$-divergence constraints. Through the analysis of gradient fields (gradient ratio and loss function surface) under diverse divergence constraints, we further illustrate the advantages of different divergence constraints in the alignment process. Regarding image-text alignment, minimal differentiation is observed among the diverse divergence constraints; conversely, for human value alignment, Jensen-Shannon (JS) divergence excels, showcasing its superior performance across all four evaluation metrics. In generative diversity evaluation, we observe that diverse divergence constraints demonstrate strengths in various aspects of diversity. Furthermore, it has been observed that increased generation diversity consistently correlates with a decrease in alignment performance. After thorough comparison, we advocate for the selection of Jensen-Shannon (JS) divergence as the foremost option in practice, which is a better trade-off between alignment performance and generation diversity.
\section*{Acknowledgment}
This work is partly supported by the National Natural Science Foundation of China (No.62103225), Natural Science Foundation of Shenzhen (No.JCYJ20230807111604008), Natural Science Foundation of Guangdong Province (No.2024A1515010003) and National Key Research and Development Program (No.2022YFB4701402).
\bibliographystyle{unsrtnat}
\bibliography{reference}

\clearpage
\appendix

\section{Additional Related Work Statement}
\subsection{Reinforcement from Human Feedback (RLHF).} 
Reinforcement learning from human feedback (RLHF) \cite{christiano2017deep,ziegler2019fine} is a crucial method for aligning artificial intelligence systems with human values, ensuring that AI systems operate and make decisions in accordance with human goals. It often integrates three core components \cite{casperopen}: feedback collection, reward modeling, and policy optimization. It facilitates humans in communicating goals without the need for manually specifying a reward function. And it leverages human judgments, which are often easier to obtain than demonstrations. Furthermore, RLHF can mitigate reward hacking compared to manually specified proxies, making reward shaping more natural and implicit. Hence, RLHF has been proven to be a valuable tool for assisting policies in learning intricate solutions in control environments \cite{hejna2023few} and for fine-tuning large scale models \cite{bai2022training,blacktraining}. Despite its widespread adoption, it still faces several limitations and open problems. In the work \cite{casperopen}, they are summarized as four aspects:  challenges with obtaining human feedback; challenges with the reward model training; challenges from policy optimization and  challenges with jointly training process. Moreover, it is pointed out that several of such weaknesses can be mitigated through the enhancement of the RLHF approach; and alternatively, some of these weaknesses can be offset by implementing additional safety measures; while others requires avoiding or compensating for with non-RLHF approaches.

\subsection{Fine-tuning Large Language Models with Reinforcement Learning 
.} 
Before RLHF, LLMs are typically aligned with human preferences through supervised fine-tuning (SFT) on demonstration data. The integration of RLHF into the training process of large language models (LLMs) has marked a significant milestone to the field of foundation model development. It has enabled LLMs to achieve human-level performance on various tasks, including text summarization, machine translation, question answering, and so on. In RLHF based fine-tuning pipeline, LLMs are trained by using human feedback as reward signals, guiding the models towards generating more accurate, relevant, and informative responses. Such iterative process allows LLMs to continuously learn and improve their performance, and this paradigm has led to the emergence of numerous remarkable models, such as OpenAI's GPT-4 \cite{achiam2023gpt}, Meta's Llama 3 \cite{touvron2023llama}, Google's Gemma \cite{team2024gemma}, and so on. Prior works has used policy-gradient methods \cite{schulman2017proximal} to this end. While they are indeed quite successful, they often come with high cost training, require extensive hyperparameter tuning process \cite{ramamurthyreinforcement}, and can be vulnerable to reward hacking, as demonstrated in various studies \cite{skalse2022defining,gao2023scaling}. Recent methods have emerged that fine-tune policy models by directly training them with a ranking loss on preference data, such as direct preference optimization (DPO) \cite{rafailov2023direct}, which have been shown to achieve performance on par with RLHF.

\subsection{Denoising Diffusion Probabilistic Models.} Denoising diffusion probabilistic models (DDPMs) have become a leading force in generative modeling due to their remarkable ability to generate diverse data formats. Diffusion model class utilizes an iterative denoising process to transform Gaussian noise into samples that adhere to a learned data distribution. Initially introduced in \cite{sohl2015deep}, further develop and promote in \cite{ho2020denoising}, they have been proved to be highly effective in a range of domains, including image generation \cite{zhang2023survey}, audio generation \cite{liao2024baton}, video generation \cite{xing2023survey}, 3D synthesis \cite{chan2023generative}, robotics \cite{kapelyukh2023dall}, and so on. Diffusion models, integrating with large-scale language encoders, have demonstrated remarkable performance in text-to-image generation \cite{raffel2020exploring,radford2021learning}. Advancements in text-to-image generation diffusion models have revolutionized the creation of lifelike visual representations based on written descriptions \cite{ramesh2021zero} and such breakthrough has opened exciting opportunities in digital art and design. In order to achieve more precise control over the outputs generated by diffusion models, researchers are exploring innovative methods to guide the diffusion process. While existing text-to-image models have achieved impressive results, they still exhibit several limitations, including challenges with compositionality, attribute binding, and so on. Researchers have also conducted extensive work to improve these aspects. Adapters \cite{zhang2023adding} have been developed to impose additional input constraints, ensuring that the generated content aligns more precisely with specific standards. For the sake of enhancing image quality and generation control, compositional approaches \cite{liu2022compositional,fengtraining} have been developed to integrate multiple models effectively.

Considering the data distribution $x_{0} \sim q_{0}(x_{0}),x_{0}\in \mathbb{R}^{n}$. DDPM algorithm approximates the data distribution $q_{0}$ with a parameterized model with the form of $p_{\theta}(x_{0})= \int p_{\theta}(x_{0:T}|c) d x_{1:T}$, where  $p_{\theta}(x_{0:T}|c)$=$p_{T}(x_{T})\prod_{t=1}^{T} p_{\theta}(x_{t-1}|x_{t},c)$, and $c$ is the conditioning information, i.e., image category and image caption. Then, we can describe the reverse process to be an Markov chain with dynamics as follows:
\[
p(x_{T})=\mathcal{N}(0,\mathit{I}), p_{\theta}(x_{t-1}|x_{t},c)=\mathcal{N}(x_{t-1};\mu_{\theta}(x_{t},c),\Sigma_{t})
\]
Furthermore, DDPMs exploits an approximate posterior $q(x_{1:T}|x_{0},c)$, namely the forward process, adding Gaussian noise to the data acccording to the variance coefficients $\beta_{1},...,\beta_{T}$:
\[
q(x_{1:T}|x_{0},c)=\prod\limits_{t = 1}^{T}q(x_{t}|x_{t-1}),\]
\[q(x_{t}|x_{t-1},c)=\mathcal{N}(\sqrt{1-\beta_{t}}x_{t-1},\beta_{t}I),
\]
\[
\alpha_{t}=1-\beta_{t},\widetilde{\alpha_{t}} =\prod\limits_{i = 1}^{t}\alpha_{i},\widetilde{\beta_{t}}=\frac{1-\widetilde{\alpha}_{t-1}}{1-\widetilde{\alpha_{t}}}.
\]
Based on these, in the work \cite{ho2020denoising}, parameterization is applied as follows:
\[
\mu_{\theta}(x_{t},c)=\frac{1}{\sqrt{\alpha_{t}}}(x_{t}-\frac{\beta_{t}}{\sqrt{1-\widetilde{\alpha_{t}}}}\epsilon_{\theta}(x_{t},c))
\]

\subsection{Fine-tuning Text-to-Image Diffusion Models with Reinforcement Learning.} Although reinforcement learning from human feedback has been widely used to align large language models, its application to diffusion models remains largely unexplored. Reward-weighted likelihood maximization \cite{lee2023aligning} proposes a three-stage fine-tuning method that leverages RLHF to enhance the alignment of text-to-image models. Rather than utilizing the reward model in dataset construction process, it is leveraged for the coefficients of loss function. DOODL \cite{wallace2023end} optimizes the initial diffusion noise vectors with respect to the loss on images generated from the full-chain diffusion, meaning that it improves a single generation iteratively at inference time. DRAFT \cite{clarkdirectly} proposes a simple and effective method for fine-tuning generative models to maximize differentiable reward functions. ReFL \cite{xu2024imagereward} utilizes a two-stage approach for diffusion model fine-tuning. In the first stage, leveraging human preference data, a reward model named ImageReward is trained, which is used for guiding the subsequent fine-tuning process. During fine-tuning, ReFL randomly selects timesteps to predict the final image with the purpose of stabilizing the training process and preventing it from focusing solely on the last step. DDPO \cite{blacktraining}   proposes a reinforcement learning (RL) framework for fine-tuning diffusion models. By defining the denoising process of diffusion models as a MDP problem, it update the pre-trained model with policy gradients to maximize the feedback-trained reward. DPOK \cite{fan2024reinforcement} is also a RL-based approach to similarly maximize the scored reward; furthermore, DPOK integrates policy optimization with reverse KL regularization for both RL fine-tuning and supervised fine-tuning.

\section{Typical DPO-based Text-to-Image Diffusion Alignment}

Advancing from the significant accomplishments of Direct Preference Optimization (DPO) in alignment, previous researches have explored its application in the application of text-to-image diffusion models, particularly Diffusion-DPO and D3PO, whose efforts established robust paradigms.

\subsection{Diffusion-DPO.} Diffusion-DPO \cite{wallace2024diffusion} offers an enhanced solution to the text-to-image alignment problem by leveraging the DPO algorithm, which is initially proposed for LLM alignment. We can begin by implementing the following symbol conversions:
\begin{itemize}
    \item The input question to the text input: $x \rightarrow c$ ;
    \item The output answer to the generated image: $y \rightarrow x_0$ ;
    \item The policy of large language models to the sampling probability of diffusion models: $\pi(y|x) \rightarrow p(x_0|c)$ ; 
    \item The preference data for output answers to the preference data for generated images: $(x,y_w,y_l)\rightarrow (c,x_0^w,x_0^l)$.
\end{itemize}
Given the settings, our goal is to optimize $p(x_0|c)$. However, when it comes to applying DPO, challenges are presented particularly for the sake that calculating sampling probability $p(x_0|c)$ requires integration over the whole sampling path $(x_1, x_2, ..., x_T)$ and $p(x_0|c)$ therefore is not computable. Consequently, it modifies the objective into optimizing the distribution of the sampling paths.
Based on this, a reinforcement learning (RL)-based objective function is formulated as follows: 
\begin{align}
\arg \max \limits_{p_{\theta}}\mathbb{E}_{c\sim \mathcal{D}_c,x_{0:T}\sim p_{\theta}(x_{0:T}|c)}\left[r(c,x_0)\right]
-\beta \mathbb{D}_{\text{KL}}\left[p_{\theta}(x_{0:T}|c)||p_{ref}(x_{0:T}|c) \right]
\end{align}
Furthermore, the loss function for Diffusion-DPO can be derived as follows:
\begin{align}
\mathcal{L}_{\text{\tiny{Diffusion-DPO}}}(\theta)=-\mathbb{E}_{\substack{(c,x_0^{w},x_0^{l})\sim \mathcal{D},\\x_{1:T}^{w}\sim p_{\theta}(x_{1:T}^{w}|x_0^w,c),\\x_{1:T}^{l}\sim p_{\theta}(x_{1:T}^{l}|x_0^l,c)}} 
\log \sigma\left(\beta \log \frac{p_{\theta}(x_{0:T}^{w}|c)}{p_{\text{ref}}(x_{0:T}^{w}|c)}-\beta \log \frac{p_{\theta}(x_{0:T}^{l}|c)}{p_{\text{ref}}(x_{0:T}^{l}|c)} \right)
\end{align}
To enhance the training efficiency, Diffusion-DPO utilizes Jensen's inequality and the convexity of the function $-\log \sigma$ to optimize an upper bound of the original objective function as follows:
\begin{align}
    -\mathbb{E}_{\substack{(c,x_0^w,x_0^l)\sim \mathcal{D},t\sim \mathcal{U}(0,T),\\x_{t-1,t}^{w} \sim p_{\theta}(x_{t-1,t}^{w}|x_0^w,c),\\x_{t-1,t}^{l} \sim p_{\theta}(x_{t-1,t}^{l}|x_0^l,c).}}
    \log \sigma \Bigg( \beta T \log \frac{p_{\theta}(x_{t-1}^{w}|x_t^w,c)}{p_{\text{\tiny{ref}}}(x_{t-1}^{w}|x_t^w,c)}-\beta T \log \frac{p_{\theta}(x_{t-1}^{l}|x_t^l,c)}{p_\text{\tiny{ref}}(x_{t-1}^{l}|x_t^l,c)} \Bigg)
\end{align}
\subsection{Direct Preference for Denoising Diffusion Policy Optimization (D3PO).} 
D3PO \cite{yang2024using} approaches the denoising process as a multi-step Markov Decision Process (MDP) and using the following mapping relationship:
\begin{align}
\label{mapping}
\begin{split}
s_t = \left( c,t,x_{T-t} \right)&;a_t=x_{T-1-t}; \\ P(s_{t+1}|s_t,a_t)=({\delta_c,\delta_{t+1},\delta_{x_{T-1-t}}})&;   \rho_0(s_0)=(p(c),\delta_0,\mathcal{N}(0,I)); \\
\pi(a_t|s_t)=p_{\theta}(x_{T-1-t}|x_{T-t},c)&;
r(s_t,a_t)=r((c,t,x_{T-t}),x_{T-t-1})
\end{split}
\end{align}
where $\delta_x$ represents the Dirac delta distribution, and $T$ denotes the maximize denoising timesteps. It sets up a kind of sparse reward: $\forall s_t,a_t $, $r(s_t,a_t) = 1$ for preferred, while $r(s_t,a_t) = -1$ for dispreferred. 

Furthermore, D3PO posits that preference for one segment implies that all state-action pairs within the segment are considered superior to those in the other segment. Under such assumption, $T$ sub-segments can be conducted for the alignment process efficiently:
\[
\sigma_{i}=\{s_i,a_i,s_{i+1},a_{i+1},...,s_{T-1},a_{T-1} \} \quad 0\leq i \leq T-1
\]
And the overall loss of D3PO algorithm can be calculated with these sub-segments as follows:
\begin{align}
\mathcal{L}_{i}(\theta)=-\mathbb{E}_{(s_i,\sigma_w^i,\sigma_l^i)} \log \rho \Big( \beta \log \frac{\pi_{\theta}(a_i^w|s_i^w)}{\pi_{\text{ref}}(a_i^w|s_i^w)} 
 - \beta \log \frac{\pi_{\theta}(a_i^l|s_i^l)}{\pi_{\text{ref}}(a_i^l|s_i^l)}
\Big)
\end{align}
where $i \in [0,T-1]$ ; $\sigma_w^i=\{ s_i^w,a_i^w,...,s_{T-1}^w,a_{T-1}^w \}$ denotes the segment preferred over the other segment $\sigma_l^i=\{ s_i^l,a_i^l,...,s_{T-1}^l,a_{T-1}^l \}$.

\subsection{Step-aware Preference Optimization (SPO).}  Contrary to the prevailing assumption that a uniform preference ordering across all stages of the diffusion process aligns with the final output images, Step-aware Preference Optimization (SPO) posits that this assumption fails to account for the nuanced effectiveness of denoising at each individual stage. SPO addresses such limitation by employing a step-aware preference model and a step-wise resampler. At the $t$-th denoising timestep, a small set $\{x_{t-1}^1,x_{t-1}^2,...,x_{t-1}^{k} \}$ is sampled, from which a preference pair $(x_{t-1}^{w},x_{t-1}^{l})$ is established by selecting the most preferred item $x_{t-1}^w$ and the most dispreferred one $x_{t-1}^{l}$. A set of preference pairs can be obtained at the $t$-th timestep by sampling from various prompts. And the DPO loss at the $t$-th timestep can be expressed as follows:
\begin{align}
\mathcal{L}_{t}(\theta)=-\mathbb{E}_{(x_{t-1}^{w},x_{t-1}^{l})\sim p_{\theta}(x_{t-1}|c,t,x_t)} \log \sigma \Bigg( \beta \log \frac{p_{\theta}(x_{t-1}^w|c,t,x_t)}{p_{\text{\tiny{ref}}}(x_{t-1}^w|c,t,x_t)} - \beta \log \frac{p_{\theta}(x_{t-1}^l|c,t,x_t)}{p_{\text{\tiny{ref}}}(x_{t-1}^l|c,t,x_t)}\Bigg)
\end{align}
where $c$ refers to the prompt and $p(c)$ is the distribution of the prompts.\\
Furthermore, the final SPO objective for all $T$ timesteps can be derived as:
\begin{align}
\begin{split}
\mathcal{L}(\theta)=-\mathbb{E}&_{t\sim \mathcal{U}[1,T],c\sim p(c),x_T \sim \mathcal{N}(0,I)(x_{t-1}^{w},x_{t-1}^{l})\sim p_{\theta}(x_{t-1}|c,t,x_t)}\\ &\quad \quad \quad \quad \quad \quad  \quad \quad 
\log \sigma \Bigg( \beta \log \frac{p_{\theta}(x_{t-1}^w|c,t,x_t)}{p_{\text{\tiny{ref}}}(x_{t-1}^w|c,t,x_t)} - \beta \log \frac{p_{\theta}(x_{t-1}^l|c,t,x_t)}{p_{\text{\tiny{ref}}}(x_{t-1}^l|c,t,x_t)}\Bigg)
\end{split}
\end{align}

\section{Detailed Mathematical Derivation}
\setcounter{theorem}{0}
\setcounter{lemma}{0}
\setcounter{fact}{0}
\setcounter{corollary}{0}
\setcounter{definition}{0}
\setcounter{remark}{0}
\setcounter{example}{0}
\setcounter{assumption}{0}
\setcounter{question}{0}
In this section, we will provide detailed proofs of Theorems.
\begin{theorem}
\label{f-Diffusion-DPO-appendix}
If $p_{\mathrm{ref}}(x_{0:T}|c) > 0$ holds for all condition $c$, $f'(x)$ is an invertible function and $0$ is not in definition domain of function $f'(x)$, the reward class consistent with Bradley-Terrry model can be reparameterized with the policy preference $p_{\theta}(x_{0:T})$ and the reference preference $p_{\mathrm{ref}}(x_{0:T}|c)$ as:
\begin{align}
\label{r(c,x_{0:T})-appendix}
    r(c,x_0)=\beta f'\Big(\frac{p_{\theta}(x_{0:T})}{p_{\mathrm{ref}}(x_{0:T}|c)}\Big) + \mathrm{const}
\end{align}
\end{theorem}
\proof
Consider the following optimal problem:
\begin{align}
\begin{split}
\max \limits_{p_{\theta}} \mathbb{E}_{c\sim p_c,x_{0:T}\sim p_{\theta}(x_{0:T}|c)}[r(c,x_{0:T})] &-\beta D_f\Big(p_{\theta}(x_{0:T}|c),p_{\text{ref}}(x_{0:T}|c)\Big)\\ 
\text{s.t.} \sum \limits_{x_{0:T}}p_{\theta}(x_{0:T}|c)=1 &;  \; \forall x_{0:T} \quad
p_{\theta}(x_{0:T}|c)\geq 0
\end{split}
\end{align}
Defining the Lagrange function as:
\begin{align}
\begin{split}
\mathcal{L}(p_{\theta}(&x_{0:T}|c),\lambda,\zeta(x_{0:T}))=\mathbb{E}_{c\sim p_c,x_{0:T}\sim p_{\theta}(x_{0:T}|c)}[r(c,x_{0:T})]\\
-\beta &\mathbb{E}_{p_{\text{ref}}(x_{0:T}|c)}\Bigg[ f\Bigg( \frac{p_{\theta}(x_{0:T}|c)}{p_{\text{ref}}(x_{0:T}|c)} \Bigg) \Bigg] - \lambda (\sum \limits_{x_{0:T}}p_{\theta}(x_{0:T})-1)
+\sum \limits_{x_{0:T}} \zeta(x_{0:T})p_{\theta}(x_{0:T}|c)
\end{split}
\end{align}

We conduct the analysis using \textit{Karush-Kuhn-Tucker (KKT) condition} as follows.

Firstly, the stationarity condition necessitates that the gradient of the Lagrangian function with respect to the primal variables be equal to zero:
\[
\nabla_{p_{\theta}(x_{0:T}|c)}\mathcal{L}(p_{\theta}(x_{0:T}|c),\lambda,\zeta(x_{0:T}))=0 ;
\]

After performing the calculations, it can be determined that:
\begin{align}
r(c,x_0)-\beta f'\Bigg( \frac{p_{\theta}(x_{0:T}|c)}{p_{\text{ref}}(x_{0:T}|c)} \Bigg) -\lambda +\zeta(x_{0:T})=0
\end{align}

Hence, we can get the formula of reward class preliminarily:
\[
r(c,x_0)=\beta f'\Bigg( \frac{p_{\theta}(x_{0:T}|c)}{p_{\text{ref}}(x_{0:T}|c)} \Bigg) +\lambda -\zeta(x_{0:T})
\]

Furthermore, we would like to consider the dual feasibility, which means the Lagrange multiplier corresponding to inequality constraint must be non-negative:
\[
\forall x_{0:T}, \zeta(x_{0:T})\geq 0
\]
And the primal feasibility holds:
\[
\sum \limits_{x_{0:T}}p_{\theta}(x_{0:T}|c)=1 ; \; \forall x_{0:T} \;
p_{\theta}(x_{0:T}|c)\geq 0
\]

Finally, we would like to consider the complementary slackness, which shows the fact that the inequality constraint must either meet with equality or have Lagrange multipliers that are zero:
\begin{align}
\forall x_{0:T}, \quad  \zeta(x_{0:T})\cdot p_{\theta}(x_{0:T}|c)=0
\end{align}

Since we have made the assumption that 0 is not in the definition domain of function $f'(x)$, which shows the fact that $\frac{p_{\theta}(x_{0:T}|c)}{p_{\text{ref}}(x_{0:T}|c)}>0$ always holds true. Moreover, we have assumed that  $p_{\text{ref}}(x_{0:T}|c) >0$ holds for all condition $c$; hence, we can draw the conclusion that $p_{\theta}(x_{0:T}|c)>0$  always holds true. Therefore, we must have:
\[
\forall x_{0:T}; \zeta(x_{0:T})=0
\]
The formula of reward class can be written as:
\[
r(c,x_0)=\beta f'\Bigg( \frac{p_{\theta}(x_{0:T}|c)}{p_{\text{ref}}(x_{0:T}|c)} \Bigg) +\lambda 
\]

The constant $\lambda$ in the formula is independent of $x_{0:T}$, which could be canceled out when applying into the Bradley-Terry model. So far, we have completed the proof.
\endproof

\begin{theorem}
\label{f-Diffusion-DPO-corollary-appendix}
In the substitution process of Bradley-Terry model, the constant
 $ \lambda $ is independent of
$x_{0:T}$ and thus can be canceled out, resulting in the following form:

\begin{align}
\mathcal{L}(\theta)=\mathbb{E}_{\substack{(c,x_0^{w},x_0^{l})\sim \mathcal{D},\\x_{1:T}^{w}\sim p_{\theta}(x_{1:T}^{w}|x_0^w,c),\\ x_{1:T}^{l}\sim p_{\theta}(x_{1:T}^{l}|x_0^l,c).}} -\log \sigma\left[\beta f' \left(\frac{p_{\theta}(x_{0:T}^{w}|c)}{p_{\text{ref}}(x_{0:T}^{w}|c)}\right)-\beta f' \left( \frac{p_{\theta}(x_{0:T}^{l}|c)}{p_{\text{ref}}(x_{0:T}^{l}|c)} \right)\right]
\end{align}
where $\sigma(\cdot)$ is the Sigmoid function; $f'(\cdot)$ represents the derivatives of $f(\cdot)$; $\beta$ is the penalty coefficient.
\end{theorem}
\proof
We know that the Bradley-Terry (BT) model provides a framework for representing human preferences as a function of pairwise comparisons:
\[
p_{\text{BT}}(x_{0:T}^w \succ x_0^l) = \sigma(r_{\phi}(c,x_{0:T}^w) - r_{\phi}(c,x_{0:T}^l))
\]
where $r_{\phi}(c,\cdot)$ represents reward function reparameterized by network $\phi$. Furthermore, the loss function can be written as maximum likelihood formula for binary classification:
\[
\mathcal{L}_{\text{BT}}=-\mathbb{E}_{c,x_{0:T}^w,x_{0:T}^l}[\log \sigma (r_{\phi}(c,x_{0:T}^w)-r_{\phi}(c,x_{0:T}^l))].
\]

Plugging Equation \eqref{r(c,x_{0:T})-appendix} into aforementioned loss function, canceling out the constant $\lambda$, and we can get the generalized formula: 
\[
\mathcal{L}(\theta)=\mathbb{E}_{\substack{(c,x_0^{w},x_0^{l})\sim \mathcal{D},\\x_{1:T}^{w}\sim p_{\theta}(x_{1:T}^{w}|x_0^w,c),\\ x_{1:T}^{l}\sim p_{\theta}(x_{1:T}^{l}|x_0^l,c)}}
-\log \sigma\left[\beta f' \left(\frac{p_{\theta}(x_{0:T}^{w}|c)}{p_{\text{ref}}(x_{0:T}^{w}|c)}\right)-\beta f' \left( \frac{p_{\theta}(x_{0:T}^{l}|c)}{p_{\text{ref}}(x_{0:T}^{l}|c)} \right)\right]
\]
\endproof
Concentrating instead on a more general formulation of the loss function as follows:
\begin{align}
\label{generalized loss function_appendix}
    \mathcal{L}_f\left( \mathrm{X}_1, \mathrm{X}_2 \right)=-\mathbb{E}\Big[\log \sigma \big( \beta f'(\mathrm{X}_1)-\beta f'(\mathrm{X}_2) \big)\Big],
\end{align}
where $\mathrm{X_1}$ is the training win ratio, and is equivalent to $\frac{p_{\theta}(x_{0:T}^{w}|c)}{p_{\text{ref}}(x_{0:T}^{w}|c)}$; similarly, $\mathrm{X}_2$ is the training loss ratio, and is identical to $\frac{p_{\theta}(x_{0:T}^{l}|c)}{p_{\text{ref}}(x_{0:T}^{l}|c)}$. 

\begin{theorem}
    The partial derivatives (gradients) of $\mathrm{X_1}$ and $\mathrm{X_2}$ resulting from Equation \eqref{generalized loss function} can be expressed as follows:
\[
\left\{
\begin{aligned}
\frac{\partial \mathcal{L}_f(\mathrm{X_1},\mathrm{X_2})}{\partial \mathrm{X_1}}=-\beta \left(1-\sigma \left( 
\beta f'(\mathrm{X}_1)-\beta f'(\mathrm{X}_2) \right)\right)f''(\mathrm{X}_1)\\
\frac{\partial \mathcal{L}_f(\mathrm{X_1},\mathrm{X_2})}{\partial \mathrm{X_2}}=\beta \left(1-\sigma \left( 
\beta f'(\mathrm{X}_1)-\beta f'(\mathrm{X}_2) \right)\right)f''(\mathrm{X}_2)
\end{aligned}
\right.
\]
Thus, the gradient ratio of $\mathcal{L}_f(\mathrm{X_1},\mathrm{X_2})$ between enhancement in probability for human-preferred responses ($\mathrm{X}_1$) and reduction in probability for human-dispreferred responses ($\mathrm{X}_2$) has the expression:
\begin{align}
    \label{gradient ratio_appendix}
        \Bigg|  \frac{\partial \mathcal{L}_f(\mathrm{X_1},\mathrm{X_2})}{\partial \mathrm{X_1}} /  \frac{\partial \mathcal{L}_f(\mathrm{X_1},\mathrm{X_2})}{\partial \mathrm{X_2}} \Bigg| = \frac{f''(\mathrm{X}_1)}{f''(\mathrm{X}_2)}
\end{align}
\label{gradients_divergence_appendix}
\end{theorem}
\proof
It is known that derivative of sigmoid function is given by the following equation:
 \[
 \sigma (x)' = \sigma (x) \cdot (1-\sigma(x))
 \]
Hence,

\[
\begin{split}
&\frac{\partial \mathcal{L}_f(\mathrm{X_1},\mathrm{X_2})}{\partial \mathrm{X_1}} \\ =&- \frac{1}{\sigma \left( \beta f'(\mathrm{X}_1)-\beta f'(\mathrm{X}_2) \right)}\cdot \sigma \left( \beta f'(\mathrm{X}_1)-\beta f'(\mathrm{X}_2) \right) \cdot (1-\sigma \left( \beta f'(\mathrm{X}_1)-\beta f'(\mathrm{X}_2) \right))\cdot \beta f''(\mathrm{X}_1)\\
=&- (1-\sigma \left( \beta f'(\mathrm{X}_1)-\beta f'(\mathrm{X}_2) \right))\cdot \beta f''(\mathrm{X}_1)\\
=&-\beta (1-\sigma \left( \beta f'(\mathrm{X}_1)-\beta f'(\mathrm{X}_2) \right)) f''(\mathrm{X}_1)
\end{split}
\]

\[
\begin{split}
&\frac{\partial \mathcal{L}_f(\mathrm{X_1},\mathrm{X_2})}{\partial \mathrm{X_2}} \\ =&\frac{1}{\sigma \left( \beta f'(\mathrm{X}_1)-\beta f'(\mathrm{X}_2) \right)}\cdot \sigma \left( \beta f'(\mathrm{X}_1)-\beta f'(\mathrm{X}_2) \right) \cdot (1-\sigma \left( \beta f'(\mathrm{X}_1)-\beta f'(\mathrm{X}_2) \right))\cdot \beta f''(\mathrm{X}_2)\\
=&(1-\sigma \left( \beta f'(\mathrm{X}_1)-\beta f'(\mathrm{X}_2) \right))\cdot \beta f''(\mathrm{X}_2)\\
=&\beta (1-\sigma \left( \beta f'(\mathrm{X}_1)-\beta f'(\mathrm{X}_2) \right)) f''(\mathrm{X}_2)
\end{split}
\]

Thus,
\[
\Bigg|  \frac{\partial \mathcal{L}_f(\mathrm{X_1},\mathrm{X_2})}{\partial \mathrm{X_1}} /  \frac{\partial \mathcal{L}_f(\mathrm{X_1},\mathrm{X_2})}{\partial \mathrm{X_2}} \Bigg| = \Bigg| \frac{-\beta \left(1-\sigma \left( 
\beta f'(\mathrm{X}_1)-\beta f'(\mathrm{X}_2) \right)\right)f''(\mathrm{X}_1)}{\beta \left(1-\sigma \left( 
\beta f'(\mathrm{X}_1)-\beta f'(\mathrm{X}_2) \right)\right)f''(\mathrm{X}_2)}\Bigg| = \frac{f''(\mathrm{X}_1)}{f''(\mathrm{X}_2)}
\]
which completes the proof.
\endproof
\begin{remark}
\label{divergence-remark}
    If the divergence is \textbf{Reverse KL divergence}, the aforementioned equation \eqref{gradient ratio} transforms into:
    \[
    \Bigg|  \frac{\partial \mathcal{L}_f(\mathrm{X_1},\mathrm{X_2})}{\partial \mathrm{X_1}} /  \frac{\partial \mathcal{L}_f(\mathrm{X_1},\mathrm{X_2})}{\partial \mathrm{X_2}} \Bigg| =\mathbf{ \frac{\mathrm{X}_2}{\mathrm{X}_1} }
    \]
     If the divergence is \textbf{Jensen-Shannon divergence}, the aforementioned equation \eqref{gradient ratio} transforms into:
    \[
    \Bigg|  \frac{\partial \mathcal{L}_f(\mathrm{X_1},\mathrm{X_2})}{\partial \mathrm{X_1}} /  \frac{\partial \mathcal{L}_f(\mathrm{X_1},\mathrm{X_2})}{\partial \mathrm{X_2}} \Bigg| = \mathbf{\frac{\mathrm{X}_2\cdot (\mathrm{X}_2+1)}{\mathrm{X}_1 \cdot (\mathrm{X}_1+1)}}
    \]
    If the divergence is $\mathbf{\alpha}$\textbf{-divergence}, the aforementioned equation \eqref{gradient ratio} transforms into:
    \[
    \Bigg|  \frac{\partial \mathcal{L}_f(\mathrm{X_1},\mathrm{X_2})}{\partial \mathrm{X_1}} /  \frac{\partial \mathcal{L}_f(\mathrm{X_1},\mathrm{X_2})}{\partial \mathrm{X_2}} \Bigg| = \mathbf{\frac{\mathrm{X}_2^{1+\alpha}}{\mathrm{X}_1^{1+\alpha}}}
    \]
    If the divergence is \textbf{forward KL divergence}, the aforementioned equation \eqref{gradient ratio} transforms into:
    \[
    \Bigg|  \frac{\partial \mathcal{L}_f(\mathrm{X_1},\mathrm{X_2})}{\partial \mathrm{X_1}} /  \frac{\partial \mathcal{L}_f(\mathrm{X_1},\mathrm{X_2})}{\partial \mathrm{X_2}} \Bigg| = \mathbf{\frac{\mathrm{X}_2^{2}}{\mathrm{X}_1^{2}}}
    \]
\end{remark}
    \begin{fact}
    For any pairwise preference data, $\mathrm{X}_2 / \mathrm{X}_1 < 1$  always holds. As optimization advances, the value of $\mathrm{X}_1$ tends to increase to more than 1, whereas $\mathrm{X}_2$ tends to decrease to less than 1. 
\end{fact}
\begin{theorem}
\label{graident_inequality}
    As optimization progresses, we have $\mathrm{X}_2 / \mathrm{X}_1 < 1$. Hence,
    \[
    0 < \frac{\mathrm{X}_2^2}{\mathrm{X}_1^2} < \frac{\mathrm{X}_2 \cdot (\mathrm{X}_2+1)}{\mathrm{X}_1 \cdot (\mathrm{X}_1+1)}<\frac{\mathrm{X}_2}{\mathrm{X}_1} < 1 \, \; \, \, \text{and} \, \; \, \, 0 < \frac{\mathrm{X}_2^2}{\mathrm{X}_1^2}  < \frac{\mathrm{X}_2^{1.8}}{\mathrm{X}_1^{1.8}}  < \frac{\mathrm{X}_2^{1.6}}{\mathrm{X}_1^{1.6}}  < \frac{\mathrm{X}_2^{1.4}}{\mathrm{X}_1^{1.4}}  < \frac{\mathrm{X}_2^{1.2}}{\mathrm{X}_1^{1.2}}  < \frac{\mathrm{X}_2}{\mathrm{X}_1} < 1
    \]
\end{theorem}
\proof
Setting $g(x)=a^x$, where $0<a<1$, and we have $g'(x)=ln a \cdot a^x \; <0$ always holds.

Thus, $g(x)$ is a monotone decreasing function, and we can easily derive that 
\[
 0 < \left(\frac{\mathrm{X}_2}{\mathrm{X}_1}\right)^2< \left(\frac{\mathrm{X}_2}{\mathrm{X}_1}\right)^{1.8}< \left(\frac{\mathrm{X}_2}{\mathrm{X}_1}\right)^{1.6}< \left(\frac{\mathrm{X}_2}{\mathrm{X}_1}\right)^{1.4}< \left(\frac{\mathrm{X}_2}{\mathrm{X}_1}\right)^{1.2}< \frac{\mathrm{X}_2}{\mathrm{X}_1} < 1
\]
Furthermore, we know that $\mathrm{X}_2 / \mathrm{X}_1 < 1$, i.e. $\mathrm{X}_2 < \mathrm{X}_1$, and then $(\mathrm{X}_2 + 1)/(\mathrm{X}_1 + 1) < 1$; therefore,
\[
    0 < \frac{\mathrm{X}_2^2}{\mathrm{X}_1^2} < \frac{\mathrm{X}_2 \cdot (\mathrm{X}_2+1)}{\mathrm{X}_1 \cdot (\mathrm{X}_1+1)}<\frac{\mathrm{X}_2}{\mathrm{X}_1} < 1
\]
\endproof

\section{Alternate Derivation: From the Rinforcement Learning Perspective}In this section, we aim to further derive the generalized formula under f-divergence from a reinforcement learning perspective. 
Here, we adopt the premise and setup of D3PO \cite{yang2024using}, which regarding the process as a multi-step Markov Decision Process (MDP) and using the following mapping relationship:
\[
\begin{split}
s_t = \left( c,t,x_{T-t} \right)&;a_t=x_{T-1-t}; \\ P(s_{t+1}|s_t,a_t)=({\delta_c,\delta_{t+1},\delta_{x_{T-1-t}}})&;   \rho_0(s_0)=(p(c),\delta_0,\mathcal{N}(0,I)); \\
\pi(a_t|s_t)=p_{\theta}(x_{T-1-t}|x_{T-t},c)&;
r(s_t,a_t)=r((c,t,x_{T-t}),x_{T-t-1})
\end{split}
\]
where $\delta_x$ represents the Dirac delta distribution, and $T$ denotes the maximize denoising timesteps. It sets up a kind of sparse reward: $\forall s_t,a_t $, $r(s_t,a_t) = 1$ for preferred, while $r(s_t,a_t) = -1$ for dispreferred.
\begin{theorem}
\label{f-D3PO-theorem-appendix}
If $\pi_{\text{ref}}(a|s)$ holds for all $s$, $f'(x)$ is an invertible function and $0$ is not in the definition domain of function $f'(x)$, the optimal policy $\pi^{*}(a|s)$ has the expression of:
\[
\pi^{*}(a|s)=\pi_{\text{ref}}(a|s)\cdot (f')^{-1}\Bigg(\frac{Q^{*}(s,a) - \lambda}{\beta}\Bigg)
\]
\noindent where $(f')^{-1}$ is the inverse function of the derivative of function $f(x)$; $\lambda$ is a fixed, constant term that is independent of $a$. 
\end{theorem}
\proof

Consider the following optimal problem:
\[
\begin{split}
\max \limits_{\pi} \mathbb{E}_{s\sim d^{\pi},a\sim \pi(\cdot|s)}\left[ Q^{*}(s,a) \right]
-\beta D_f \left[ \pi(a|s),\pi_{\text{ref}}(a|s) \right]
\end{split}
\]
\[
s.t. \sum \limits_{a} \pi(a|s)=1 ; \forall a \,  \pi(a|s) \geq 0
\]
where $Q^{*}(s,a)$ is the optimal action-value function; $d^{\pi}=(1-\gamma)\sum_{t=0}^{\infty}\gamma^t P_t^{\pi}(s)$ represents the state visitation distribution; $s,a,\pi,P_t^{\pi}$ adhere to the definitions outlined in equation \eqref{mapping}.\\
Defining the following Lagrange function:
\begin{align}
\begin{split}
\mathcal{L}(\pi(a|s),\lambda,\xi(a))=\mathbb{E}_{s\sim d^{\pi},a\sim \pi(\cdot|s)}&\left[Q^{*}(s,a)\right]\\
- \beta \mathbb{E}_{\pi_{\text{ref}}(a|s)} \Bigg[&f\left(\frac{\pi(a|s)}{\pi_{\text{{ref}}}(a|s)}\right)  \Bigg]-\lambda \left( \sum \limits_{a}\pi(a|s) - 1 \right)
+ \xi(a) \pi(a|s)
\end{split}
\end{align}
Employing the \textit{Karush-Kuhn-Tucker (KKT) conditions} for analysis:\\
Firstly, the stationarity condition necessitates that the gradient of the Lagrangian function with respect to the primal variables should be zero:
\[
 \nabla_{\pi(a|s)}\mathcal{L}(\pi(a|s),\lambda,\xi(a))=0
\]
Performing the calculation, we can get:
\[
Q^{*}(s,a)-\beta f'\left( \frac{\pi(a|s)}{\pi_{\text{ref}}(a|s)} \right)= \lambda-\xi(a)
\]
Furthermore, considering the dual feasibility, which stipulates that the Lagrange multiplier associated with an inequality constraint must adhere to a non-negative condition:
\[
\forall  a, \, \xi(a)\geq 0
\]
And the primal feasibility shows that:
\[
\sum \limits_{a} \pi(a|s)=1 ; \forall a \,  \pi(a|s) \geq 0
\]
Moreover, thinking about the complementary slackness, which dictates that for an inequality constraint, either the constraint must be satisfied with equality, or its corresponding Lagrange multiplier must be zero:
\[
\forall a; \, \pi(a|s)\cdot \xi(a)=0
\]
Given that 0 is not in the definition domain of $f'(x)$, it follows the fact that $\frac{\pi(a|s)}{\pi_{\text{ref}(a|s)}}>0$ always holds. Moreover, we have the assumption that $\pi_{\text{ref}}(a|s)>0$ is satisfied. Hence, there must be $\pi(a|s)>0$. From the analysis that has been conducted, the subsequent conclusion is attainable:
\[
\forall a; \; \xi(a)=0
\]
Substituting the above conclusion into the stationarity condition yields:
\[
f'\left( \frac{\pi(a|s)}{\pi_{\text{ref}}(a|s)} \right)= \frac{Q^{*}(s,a) - \lambda}{\beta}
\]
Through certain algebraic computation, we can derive:
\[
\pi^{*}(a|s)=\pi_{\text{ref}}(a|s)\cdot (f')^{-1}\Bigg(\frac{Q^{*}(s,a) - \lambda}{\beta}\Bigg)
\]
So far, we have completed the proof.
\endproof
\begin{remark}
\label{Q^{*}(s,a)}
 Rearranging the equation in Theorem \ref{f-D3PO-theorem-appendix}, we can obtain the following formula:
 \[
 Q^{*}(s,a)= \beta f'\left( \frac{\pi(a|s)}{\pi_{\text{ref}}(a|s)} \right) + \lambda
 \]
\end{remark}
Substituting the result we obtained in Remark \ref{Q^{*}(s,a)} into the Bradley-Terry model \cite{19ff28b9-64f9-3656-ba40-08326a05748e}, we can similarly eliminate the constant $\lambda$ and gain the final generalized formula.

\section{Further Analysis on the Gradient Fields}

In the work \cite{yan20243d}, it is shown that the original DPO (alignment of LLMs) increasingly loses its ability to steer the direction of response optimization in LLM alignment as the alignment process advances; in other words, it risks degenerating into a mechanism that merely learns the rejected responses, rather than actively shaping the chosen responses' trajectory towards alignment. We would like to further investigate whether such phenomena continued exist in the context of text-to-image generation alignment with diverse divergence constraints.

\begin{remark}
    If the divergence is \textbf{Reverse KL divergence}, equations in Theorem \ref{gradients_divergence} can be simplified as:
\[
\label{gradients_RKL}
\left\{
\begin{aligned}
\frac{\partial \mathcal{L}_f\left( \mathrm{X_1},\mathrm{X_2} \right)}{\partial \mathrm{X_1}}&=-\beta \frac{\mathrm{X}_2^{\beta}}{\mathrm{X}_1\cdot \left( \mathrm{X_1}^{\beta} + \mathrm{X_2}^{\beta} \right)}\\
\frac{\partial \mathcal{L}_f\left( \mathrm{X_1},\mathrm{X_2} \right)}{\partial \mathrm{X_2}}&=\beta \frac{\mathrm{X}_2^{\beta-1}}{\left( \mathrm{X_1}^{\beta} + \mathrm{X_2}^{\beta} \right)}
\end{aligned}
\right.
\]
\end{remark}

\noindent The main cause of the aforementioned described phenomenon, known as \textit{model learning degradation}, occurs as $\mathrm{X}_2 \rightarrow 0$ and $\beta < 1$, resulting in $\frac{\partial \mathcal{L}_f\left( \mathrm{X_1},\mathrm{X_2} \right)}{\partial \mathrm{X_1}}$ tends 0 for the sake of $\mathrm{X}_2^{\beta} \rightarrow 0$, while $\frac{\partial \mathcal{L}_f\left( \mathrm{X_1},\mathrm{X_2} \right)}{\partial \mathrm{X_2}}$ tends infinity as a consequence of $\mathrm{X}_2^{\beta-1} \rightarrow \infty$. In fact, in our scenario, the aforementioned phenomenon is mitigated by our typical practice of assigning a relative large value to $\beta$ ($\beta=$10 in the experiments of our work), thus prevent it focus on unlearning rejected items only. 
\begin{remark}
    If the divergence is \textbf{Jensen-Shannon divergence}, equations in Theorem \ref{gradients_divergence} can be simplified as:
        \[
\label{gradients_JS}
\left\{
\begin{aligned}
\frac{\partial \mathcal{L}_f\left( \mathrm{X_1},\mathrm{X_2} \right)}{\partial \mathrm{X_1}}&=-\beta \cdot \frac{1}{\mathrm{X}_1} \cdot \frac{\mathrm{X}_2^{\beta} \left( 1+\mathrm{X}_1\right)^{\beta -1}}{\mathrm{X}_2^{\beta}(1+\mathrm{X}_1)^{\beta}+\mathrm{X}_1^{\beta}(1+\mathrm{X}_2)^{\beta}}\\
\frac{\partial \mathcal{L}_f\left( \mathrm{X_1},\mathrm{X_2} \right)}{\partial \mathrm{X_2}}&=-\beta \cdot \frac{1}{\mathrm{X}_2+1} \cdot \frac{\mathrm{X}_2^{\beta-1} \left( 1+\mathrm{X}_1\right)^{\beta }}{\mathrm{X}_2^{\beta}(1+\mathrm{X}_1)^{\beta}+\mathrm{X}_1^{\beta}(1+\mathrm{X}_2)^{\beta}}
\end{aligned}
\right.
\]
    When $\mathrm{X}_2 \rightarrow 0$, if $\beta > 1 $, both $\frac{\partial \mathcal{L}_f\left( \mathrm{X_1},\mathrm{X_2} \right)}{\partial \mathrm{X_1}} \rightarrow 0$ and $\frac{\partial \mathcal{L}_f\left( \mathrm{X_1},\mathrm{X_2} \right)}{\partial \mathrm{X_2}} \rightarrow 0$ simultaneously; conversely, if $\beta < 1$, the result would be $\frac{\partial \mathcal{L}_f\left( \mathrm{X_1},\mathrm{X_2} \right)}{\partial \mathrm{X_1}} \rightarrow 0$ but $\frac{\partial \mathcal{L}_f\left( \mathrm{X_1},\mathrm{X_2} \right)}{\partial \mathrm{X_2}} \rightarrow \infty$.
\end{remark}
\noindent Remark \ref{gradients_JS} indicates that when the \textit{Jensen-Shannon divergence} is selected as an regularization, choosing a value of $\beta$ \textit{greater than 1} is advantageous.
\begin{remark}
    If the divergence is $\mathbf{\alpha}$\textbf{-divergence}, equations in Theorem \ref{gradients_divergence} can be simplified as:
\[
\left\{
\label{gradients_alpha}
\begin{aligned}
\frac{\partial \mathcal{L}_f\left( \mathrm{X_1},\mathrm{X_2} \right)}{\partial \mathrm{X_1}}&=-\beta \cdot \frac{1}{\mathrm{X}_1^{1+\alpha}} \cdot \frac{e^{ \beta/\alpha \cdot \frac{1}{\mathrm{X}_1^{\alpha}}}}{e^{\beta/\alpha \cdot\frac{1}{\mathrm{X}_1^{\alpha}}} + e^{ 
\beta/\alpha \cdot \frac{1}{\mathrm{X}_2^{\alpha}}}}\\
\frac{\partial \mathcal{L}_f\left( \mathrm{X_1},\mathrm{X_2} \right)}{\partial \mathrm{X_2}}& = \beta \cdot \frac{1}{\mathrm{X}_2^{1+\alpha}} \cdot \frac{e^{ \beta/\alpha \cdot \frac{1}{\mathrm{X}_1^{\alpha}}}}{e^{\beta/\alpha \cdot\frac{1}{\mathrm{X}_1^{\alpha}}} + e^{ 
\beta/\alpha \cdot \frac{1}{\mathrm{X}_2^{\alpha}}}}
\end{aligned}
\right.
\]
    Given that both $\alpha$ and $\beta$ are positive values, it follows that $\frac{\partial \mathcal{L}_f\left( \mathrm{X_1},\mathrm{X_2} \right)}{\partial \mathrm{X_1}} \rightarrow 0$ and $\frac{\partial \mathcal{L}_f\left( \mathrm{X_1},\mathrm{X_2} \right)}{\partial \mathrm{X_2}} \rightarrow 0$ always hold true when $\mathrm{X}_2 \rightarrow 0$.
\end{remark}
\begin{remark}
    If the divergence is \textbf{Forward KL divergence}, equations in Theorem \ref{gradients_divergence} can be simplified as:
\[
\left\{
\label{gradients_FKL}
\begin{aligned}
\frac{\partial \mathcal{L}_f\left( \mathrm{X_1},\mathrm{X_2} \right)}{\partial \mathrm{X_1}}&=-\beta \cdot \frac{1}{\mathrm{X}_1^{2}} \cdot \frac{e^{ \frac{\beta}{\mathrm{X}_1}}}{e^{ \frac{\beta}{\mathrm{X}_1}} + e^{ 
e^{ \frac{\beta}{\mathrm{X}_2}}}}\\
\frac{\partial \mathcal{L}_f\left( \mathrm{X_1},\mathrm{X_2} \right)}{\partial \mathrm{X_2}}& = \beta \cdot \frac{1}{\mathrm{X}_2^{2}} \cdot \frac{e^{\frac{\beta}{\mathrm{X}_1}}}{e^{\frac{\beta}{\mathrm{X}_1}} + e^{ 
\frac{\beta}{\mathrm{X}_2}}}
\end{aligned}
\right.
\]
    $\forall \beta$, $\frac{\partial \mathcal{L}_f\left( \mathrm{X_1},\mathrm{X_2} \right)}{\partial \mathrm{X_1}} \rightarrow 0$ and $\frac{\partial \mathcal{L}_f\left( \mathrm{X_1},\mathrm{X_2} \right)}{\partial \mathrm{X_2}} \rightarrow 0$ always hold true when $\mathrm{X}_2 \rightarrow 0$.
\end{remark}
According to Remark \ref{gradients_alpha} and Remark \ref{gradients_FKL}, if the regularization is in terms of $\alpha$-divergence or Forward KL divergence, then no matter the value of $\beta$, it will not result in model training degradation.   

 Moreover, we would like to further discuss the relationship between generation diversity and gradient field. Within the work \cite{yang2024using}, the property of dispersion effect on unseen generations of original DPO has been proposed. It elucidates that as $\mathrm{X}_2$ rapidly decreases to 0, the gradient on $\mathrm{X}_1$ will gradually diminish, consequently leading to a stochastic decline in the likelihood of the selected response. Such dispersion effect contributes to the genration diversity. Hence, for the sake that forward KL divergence has the minimal gradient ratio and reverse KL divergence has the maximal gradient ratio, should theoretically resulting in optimal alignment diversity for forward KL divergence and poorest alignment diversity for reverse KL divergence. In fact, \cite{wangbeyond,go2023aligning} does reach such a analogous conclusion from a practical point of view in the LLM alignment task. 

However, we should also be mindful of two aspects. Firstly, in the task of LLM alignment, typically only one epoch is conducted, thus conforming well to the aforementioned theory.
Nevertheless, in the task of Text-to-Image generation alignment, multiple epochs are often performed (e.g., 10 epochs in Diffusion-DPO, SPO and experiments of our work; 1000 epochs in D3PO), which renders the diversity variations caused by the gradient ratio negligible after training for multiple epochs. Secondly, the contextual dimension of images is higher than that of text, and the evaluation indicators for image diversity often focus on different aspects of images. In our experiments, we observe that after sufficient training, $\alpha-$divergence ($\alpha$=0.6) generally achieves the best generation diversity. However, it is worth noting that while it achieves the optimal generation diversity, it performs worst in terms of human value alignment performance. And we can intuitively find that generation diversity and alignment performance are a pair of conflicting entities. To achieve the best trade-off between the two in alignment, we should first pursue better alignment performance, and then, on the basis of assured alignment performance, pursue better generation diversity. Based on a comprehensive theoretical examination and empirical evidence from experimental results, we advocate for regarding \textbf{Jensen-Shannon divergence} as the first choice in practice.

\section{Plot of Gradient Fields and Visualization of Landscapes}
\label{Plot of Gradient Fields}
In order to obtain a more intuitive understanding of the impact of different divergence choices during the alignment process, we visualize the landscape of alignment objective functions with different divergences from two viewing angles, as shown in Figure \ref{gradient_plot} (the penalty coefficient $\beta$ is selected as 10). Furthermore, to enhance intuition, we plot the gradient field of corresponding loss function on the plane where $Z$ equals 50. When it comes to consider the smoothness within loss function landscape, surface of \textit{Jensen-Shannon divergence} exhibits the best smoothness, which suggests a more stable alignment process. Moreover, this indicates a more robust alignment mechanism, which helps prevent the process from merely unlearning undesired outputs rather than actively steering chosen outputs towards optimization; and this also mitigates the phenomenon that the gradient on $\mathrm{X}_1$ gradually diminishes as $\mathrm{X}_2$ rapidly decreases to 0, which consequently leads to a stochastic decline in the likelihood of the selected response \cite{yan20243d}. 
\clearpage
\begin{figure}[H]
\centering
\includegraphics[width=0.97\columnwidth]{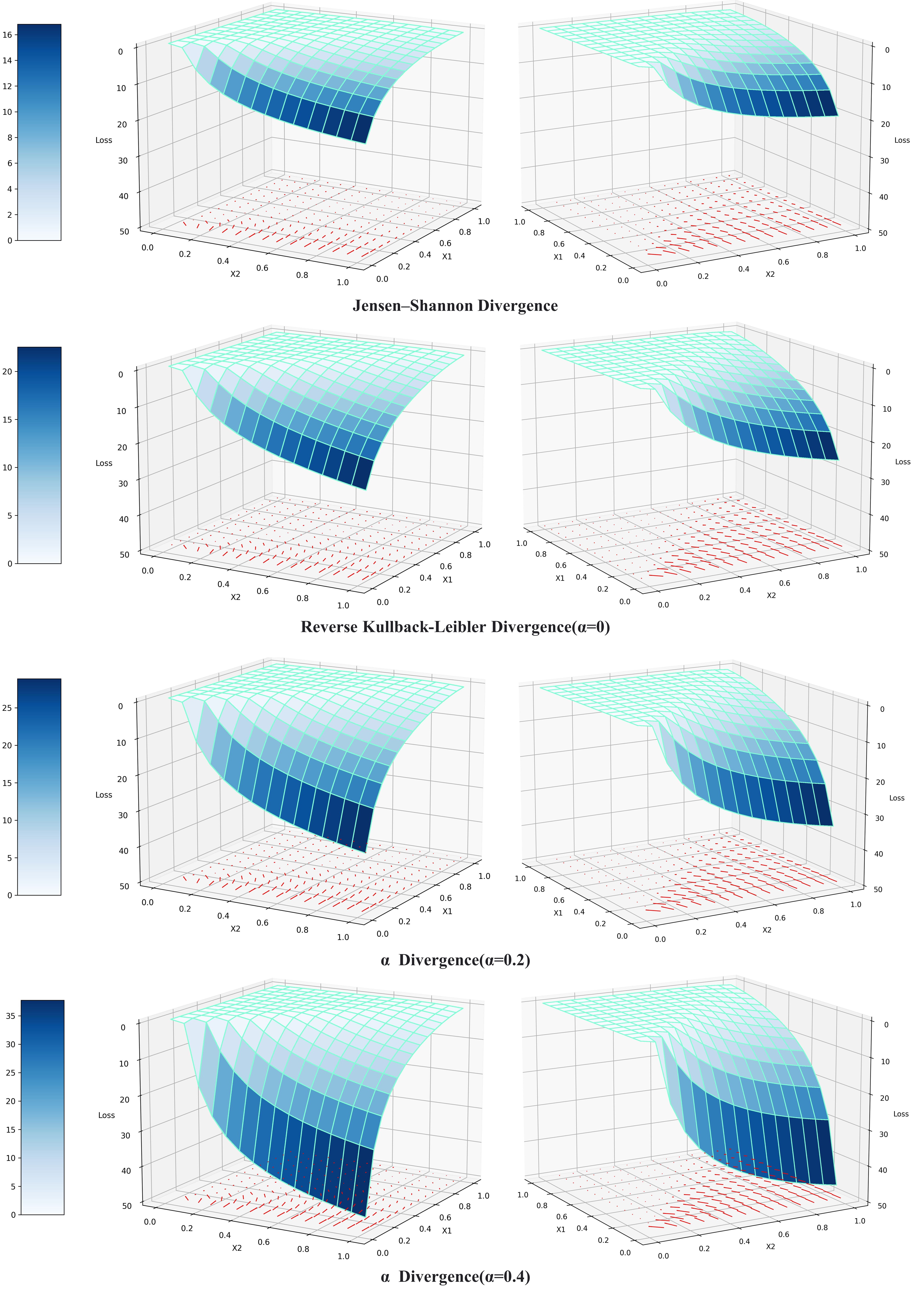}
\end{figure}
\setcounter{figure}{1}
\begin{figure}[H]
\centering
\includegraphics[width=0.97\columnwidth]{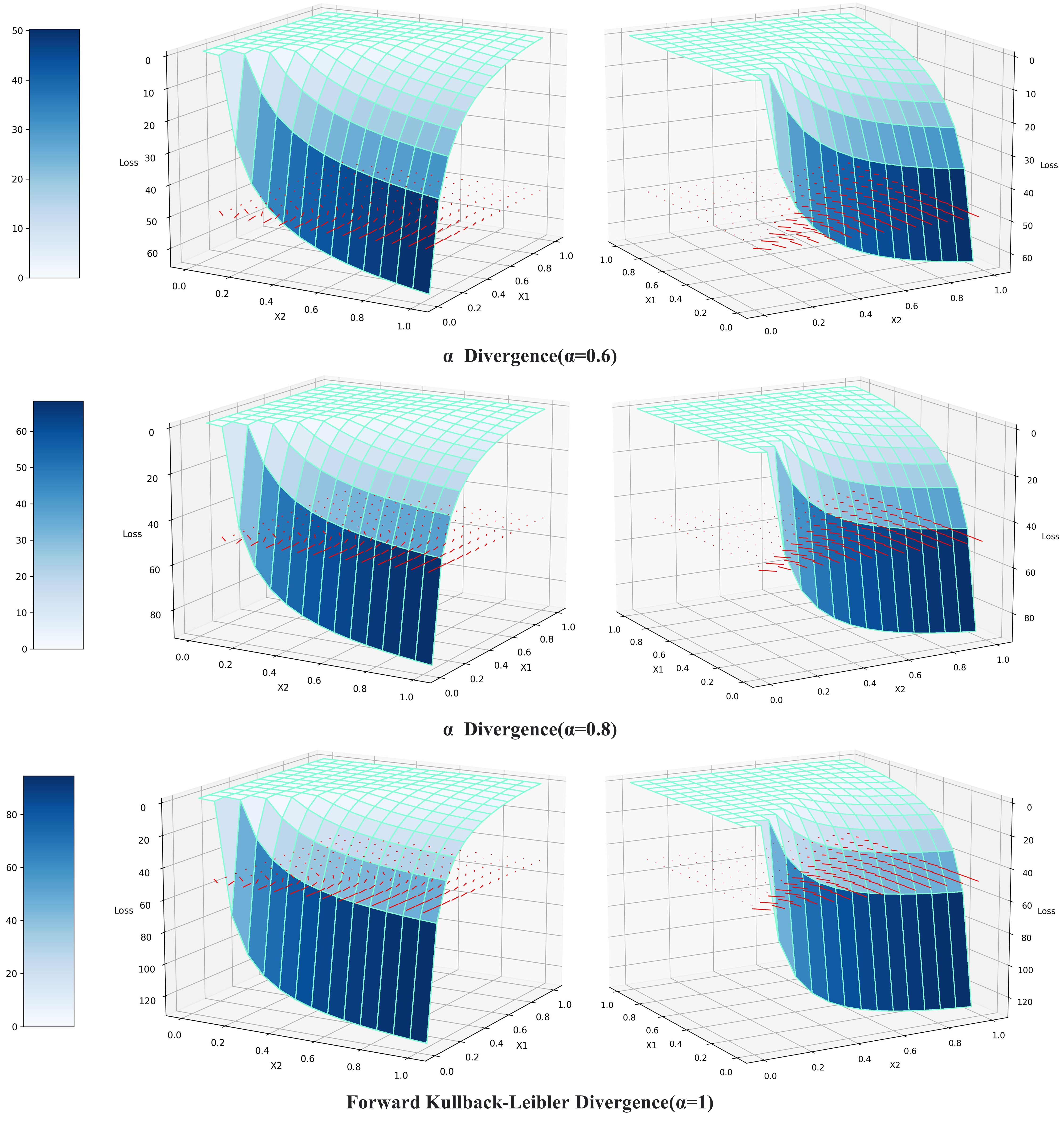}
\caption{Landscapes' visualization of alignment objective functions with different divergences from two viewing angles and gradient fields' visualization of the corresponding loss function on the plane $Z=50$.}
\label{gradient_plot}
\end{figure}

\section{Detailed Metric Description}

\subsection{Alignment Performance Metric. } In this paper, we utilize six metrics for evaluating the alignment performance. We employ the text-image CLIP score \cite{hessel2021clipscore} and VQAScore \cite{lin2025evaluating} to evaluate the performance of text-image alignment and the Aesthetics score \cite{schuhmann2022laion}, ImageReward \cite{xu2024imagereward}, PickScore \cite{kirstain2023pick} and HPS-v2 \cite{wu2023human} to evaluate the performance of human value alignment.

\textbf{Text-Image CLIP score.} The Text-Image CLIP score serves as a quantitative measure for evaluating the likeness between text-image pairs.  CLIP score is fundamentally based on the CLIP model, which transforms input text and images into distinct text and image vectors, followed by calculation of the dot product of these vectors. Foundational aim of the CLIP model is to cultivate versatile multimodal representations, free from specialized domain expertise, through the integration of linguistic indicators and visual data. Training approach of CLIP model mainly hinges on contrastive learning, where the system partitions the incoming text-image pairs into two categories: one cluster includes similar pairs to the input, whereas the other assembles dissimilar pairs. The model then learns representations of these inputs, with the objective to increment similarity within matching pairs while reducing it between non-matching pairs. Benefiting from its pre-training strategy, it enables the extraction of significant image and text features from vast unsupervised datasets. The CLIP model and CLIP score have demonstrated commendable performance across a wide range of tasks, encompassing image classification, semantic segmentation, image generation, object localization, video interpretation, and so on.

\textbf{VQAScore.}
VQAScore meticulously transforms textual cues into precise inquiries, deploying the generative vision-language models with visual-question-answering (VQA) tasks to evaluate the congruence between the image and the descriptive text. Such innovative approach streamlines the assessment process while markedly enhancing the precision and dependability of evaluations. Furthermore, by utilizing the CLIP-FlanT5 model, VQAScore fosters a reciprocal influence between the visual content and the textual query, aligning more closely with human comprehension of the interplay between the image and the text.

\textbf{Aesthetics score.} The LAION Aesthetics Predictor is utilized to estimate an image's aesthetic score, quantifying the mean human appreciation for its visual appeal.  It leverages a neural network architecture (MLP) that takes CLIP embeddings as inputs to ascertain the average preference level for the image.  Each image is assigned a score on the scale of 0 to 10, with 0 signifying the least visually attractive and 10 denoting the highest level of visual appeal.

\textbf{ImageReward.} ImageReward, leveraging a structure that combines ViT-L for image encoding and a 12-layer Transformer for text encoding, tackles the challenges of text-to-image generation to some extent, especially regarding the quality of pre-training data, which are plagued by noise and a skewed distribution that doesn't match the data users input in prompts. Notably, as a zero-shot evaluation tool, ImageReward often aligns with human judgments, demonstrating the capability to make nuanced quality comparisons between individual samples. 

\textbf{PickScore.} A comprehensive, natural dataset, dubbed "Pick-a-Pic," is compiled and utilizing the dataset, an advanced scoring function, namely "PickScore", is built. "PickScore" excels in assessing generated images against prompts, surpassing not only machine learning models but also expert human evaluations. Its utility spans multiple domains such as model evaluation, image generation enhancement, text-to-image dataset refinement, and the optimization of text-to-image models through methodologies such as Reinforcement Learning Human Feedback (RLHF). PickScore follows the architecture of CLIP; provided a prompt $x$ and an image $y$, PickScore $s$ calculates a real number through the representation of $x$ with a text encoder and
$y$ with an image encoder as two $d$-dimensional vectors, and subsequently returns their inner product:
\[
score(x,y)=\mathrm{E}_{\text{txt}}(x)\cdot\mathrm{E}_{\text{img}}(y) \cdot T
\]
where $T$ is the learned temperature parameter of CLIP.

\textbf{HPS-v2.} Human Preference Dataset v2 (HPD-v2) encapsulates human preferences for images sourced from a multitude of platforms. It consists of 798,090 individual human preference choices for 433,760 paired image comparisons. The dataset has been taken care to deliberately collect the text prompts and images to minimize potential biases, a common pitfall in previous datasets. Whereafter, through fine-tuning the CLIP model on HPD-v2, the Human Preference Score v2 (HPS-v2) is derived, a scoring model that can more accurately gauge human preferences for generated images. HPS-v2 has been shown to generalize more effectively than earlier metrics across a variety of image datasets, and it is responsive to algorithmic improvements of text-to-image generative models.

\subsection{Generation Diversity Metric} In this work, we utilize eight metrics for comprehensively evaluating generation diversity from diverse aspects: Image-Image CLIP score \cite{hessel2021clipscore}, Image Entropy (Entropy 1D and Entropy 2D) \cite{sparavigna2019entropy}, LPIPS \cite{zhang2018unreasonable}, RMSE \cite{Willmott2005AdvantagesOT}, PSNR \cite{5596999}, SSIM \cite{5596999}, FSIM \cite{zhang2011fsim}. 

\textbf{Image-Image CLIP score.} Text-Image CLIP score and Image-Image CLIP score are both grounded in the evaluation of high-dimensional embeddings produced by the CLIP model.  Similarly, the Image-Image CLIP score functions as an efficacious metric for evaluating the structural congruity between images, thereby enabling assessments of images' similarity. Hence, we select Image-Image CLIP score as an indicator for the diversity of images produced by the trained diffusion model: a diminutive CLIP score between two generated images signifies a pronounced disparity, implying that the model demonstrates a heightened capacity for generating diverse content.

\textbf{Image Entropy (Entropy 1D and Entropy 2D).} Image Entropy is a statistical metric employed to evaluate the information content and complexity within an image. It quantifies the average information per pixel, with higher entropy values signaling a greater diversity and richness in the image's information content. One-dimensional image entropy (Entropy 1D) quantifies the information encapsulated within the distribution's clustering properties of gray levels:
\[
H_{1d}=\sum \limits_{i=0}^{255}P_i \log P_i
\]
where $P_i$ presents the proportion of pixels in the image with gray level value $i$.

The one-dimensional image entropy (Entropy 1D) successfully captures the aggregation properties of gray level distribution, yet neglects spatial attributes. To rectify this discrepancy, supplementary feature metrics are incorporated, which, in conjunction with the one-dimensional entropy, serve as the cornerstone for the evolution of two-dimensional image entropy (Entropy 2D).  Such augmentation facilitates a more holistic evaluation that merges both spatial and distributional data within an image. The neighborhood gray level mean, when chosen as a spatial feature quantity in conjunction with the pixel gray levels, constitutes a feature tuple denoted as $(i, j)$:
\[
P_{i,j}=\frac{f(i,j)}{N^2}
\]
where $i$ is the gray value of the pixel, and $j$ is the mean gray value of its neighborhood; $f(i,j)$ is the occurrence frequency of characteristic binary $(i,j)$ and $N$ is the dimension of the image. Then the two-dimensional image entropy (Entropy 2D) can be defined as:
\[
H_{2d}=\sum \limits_{i=0}^{255} \sum \limits_{j=0}^{255} P_{ij} \log P_{ij}
\]

\textbf{LPIPS.}  Learned Perceptual Image Patch Similarity (LPIPS) is a deep learning-based metric designed for assessing image similarity, which is calculated based on features output by a deep convolutional neural network (AlexNet in our work). Utilizing the feature representations learned by a deep neural network, which is capable of capturing details of human visual perception such as texture, color, and structure, the computation of perceptual similarity between two images can be conducted. Firstly, pre-trained deep neural networks, notably those like AlexNet \cite{NIPS2012_c399862d} or VGG \cite{simonyan2014very}, are employed to extract features from input images. The outputs from the network's intermediate layers are then typically utilized, as they encapsulate a spectrum of abstract features, spanning from rudimentary edge and texture details to more sophisticated representations of objects and scenes, denoted as $\hat{y_1^i}$, $\hat{y_2^i}$ $ \in \mathbb{R}^{H_l\times W_l \times C_l}$. Subsequently, the distances between extracted features in the feature space can be calculated:
\[
d(y_1, y_2)=\sum \limits_l \frac{1}{H_l W_l} \sum \limits_{h,w} \| w_l \odot (\hat{y_1^i}- \hat{y_2^i}) \|_2^2
\]
and converted into a comprehensible similarity score by standardizing them to the range of 0 to 1. For our evaluation, LPIPS is selected as a metric;  a lower score indicates a higher similarity between images, while a higher score suggests greater diversity or disparity between them. 

\textbf{RMSE.}  Root Mean Square Error (RMSE) is a statistical metric that   primarily employed in statistical analysis and machine learning disciplines, serving as a benchmark for gauging the accuracy of predictions.    In our scenario, we utilize the RMSE as a criterion to measure the pixel-level differences between pairs of generated images.   Such pixel-level variance can be considered as an indicator of the images' diversity, thereby offering a quantifiable assessment of the generated images' diversity:
\[
RMSE=\sqrt{\frac{1}{L\times W}\sum \limits_{i=1}^{L\times W}(p_i-q_i)^2}
\]
where $L$ is the length of the image, $W$ is the width of the image; $p_i$ and $q_i$ are $i$-th pixels of two generated images.

\textbf{PNSR.} Peak Signal-to-Noise Ratio (PSNR), a prevalent metric in image processing and compression, quantitatively assesses the discrepancy between an original image and its modified counterpart. This metric, typically reported in decibels (dB), is characterized by a higher value indicative of a diminished difference between the two images, effectively serving as a metric for evaluating the diversity of generated images:
\[
PSNR=10 \log_{10}\Big(\frac{Max^2}{MSE}\Big)
\]
where $Max$ denotes the maximum pixel value that an image attain, $MSE$ denotes the mean squared error between two images.

\textbf{SSIM.} Structural Similarity Index Measure (SSIM) is a metric designed to assess the likeness between images by emulating the human visual system's perception of image quality. Traditional metrics (e.g. RMSE, PNSR, Image Entropy) usually focus on disparities in pixel values, whereas SSIM incorporates the structural aspects of images for evaluation. It is often executed from three aspects: luminance similarity, contrast similarity, and structural similarity - on a scale from 0 to 1:
\[
\begin{split}
l(x,y)&=\frac{2\mu_x\mu_y+c_1}{\mu_x^2+\mu_y^2+c_1}
;\\c(x,y)&=\frac{2\sigma_x\sigma_y+c_2}{\sigma_x^2+\sigma_y^2+c_2}
;\\s(x,y)&=\frac{\sigma_{xy}+c_3}{\sigma_{x}\sigma_{y}+c_3}
\end{split}
\]
where $\mu_x$ and $\mu_y$ are means of $x$ and $y$; $\sigma_x$ and $\sigma_y$ are variances of $x$ and $y$, respectively; and $\sigma_{xy}$ is the covariance of $x$ and $y$. SSIM calculates similarity between two images across these three dimensions, providing an overall similarity index ranging from 0 to 1. The closer the value is to 1, the more similar the two images are:
\[
SSIM(x,y)= \left[ l(x,y)^{\alpha}\cdot c(x,y)^{\beta}\cdot s(x,y)^{\gamma} \right]
\]
In typical situations, $\alpha$, $\beta$ and $\gamma$ are all set to 1.

\textbf{FSIM.}  Feature Similarity Index Measure (FSIM) utilizes feature similarity for assessment. The Human Visual System (HVS) bases its perception on essential visual attributes, and the phase congruency (PC) feature excels in depicting local structures. Remarkably, PC's resilience to changes in the image context guarantees the stability of feature extraction. Nonetheless, it's recognized that modifications in the image can influence visual perception. Therefore, to augment the comprehensive analysis, gradient features, particularly gradient magnitude (GM), are incorporated. Consequently, in FSIM, both PC and GM features collaborate to serve complementary roles, synergistically capturing a holistic evaluation. For two images, the calculations for $PC1$, $GM1$, $PC2$, and $GM2$ are firstly performed. Subsequently, compute the similarity for $PC$ and for $GM$ as follows:
\[
\begin{split}
S_{PC}(\mathrm{x})&=\frac{2PC_1(\mathrm{x})\cdot PC_2(\mathrm{x})+T_1}{PC_1(\mathrm{x})^2+PC_2(\mathrm{x})^2+T_1};\\
S_{GM}(\mathrm{x})&=\frac{2GM_1(\mathrm{x})\cdot GM_2(x)+T_2}{GM_1(x)^2+GM_2(x)^2+T_2}.
\end{split}
\]
Furthermore, the similarity expressed by fusion of PC and GM can be given as:
\[
S_{L}(\mathrm{x})=[S_{PC}(\mathrm{x})]^{\alpha} \cdot [S_{GM}(\mathrm{x})]^{\beta}
\]
Finally, the calculation of FSIM is described as follows:
\[
FSIM = \frac{\sum_{\mathrm{x}\in \Omega} S_L(\mathrm{x}) \cdot PC_m(\mathrm{x})}{\sum_{\mathrm{x}\in \Omega} PC_m(\mathrm{x})}
\]
\clearpage
\section{Further Larger-Scale Evaluation}

To further demonstrate the persuasiveness of our conclusions, we conduct additional evaluation on the aligned models. GenAI-Bench \cite{li2024genai}, serves as a comprehensive benchmark for compositional text-to-visual generation, and we report the evaluation results on GenAI-Bench in Table \ref{evaluations of alignment performance_Gen-AI} (alignment performance) and Table \ref{evaluations of diversity performance_Gen-AI} (diversity performance). Furthermore, we employ the parti-prompts \cite{yuscaling} training dataset and the entire HPS-v2 \cite{wu2023human} training set for evaluation, by combining the generated images from these sets together, we report the evaluation results \textit{on all 4832 prompts} in Table \ref{evaluations of alignment performance_larger-scale} (alignment performance) and Table \ref{generation diversity evaluation_larger-scale} (generation diversity). The results obtained are similar to that in the main paper. \textbf{Jensen-Shannon divergence} exhibits the best alignment performance and suboptimal generation diversity, achieving the best trade-off.

\setlength{\tabcolsep}{1.5mm}
\begin{table*}[ht]\scriptsize
\centering
\begin{tabular}{cc|cc|cccc}
\toprule
\multicolumn{2}{c|}{Model}& \multicolumn{1}{c|}{CLIPScore $\uparrow$}& \multicolumn{1}{c|}{VQAScore $\uparrow$} & \multicolumn{1}{c|}{Aesthetics Score $\uparrow$} & \multicolumn{1}{c|}{ImageReward $\uparrow$} & \multicolumn{1}{c|}{Pickscore $\uparrow$} &  {HPS-V2   $\uparrow$}                           \\ \midrule \midrule 
\multicolumn{2}{c|}{Original Model}  & 0.334$\pm$0.046 & 0.638$\pm$0.268 & 5.433$\pm$0.427                    & 0.195$\pm$0.996                   & 21.446$\pm$1.143                 & \multicolumn{1}{c}{27.148$\pm$1.463} \\ \midrule \midrule
\multicolumn{2}{c|}{Reverse KL Divergence}& \textbf{0.344$\pm$0.045} &  \textbf{0.669$\pm$0.265} & 5.582$\pm$0.408                     & 0.556$\pm$0.947                  & 21.889$\pm$1.153                 & 27.827$\pm$1.452                      \\ \midrule
\multicolumn{1}{c|}{\multirow{5.20}{*}{$\alpha$-Divergence}} & $\alpha$=0.2 & 0.344$\pm$0.046 &  0.665$\pm$0.270  & 5.607$\pm$0.428                     & 0.535$\pm$0.949                  & 21.850$\pm$1.168                 & \textit{\textbf{27.900$\pm$1.475} }                     \\ \cmidrule{2-8} 
\multicolumn{1}{c|}{}                              & $\alpha$=0.4 & 0.343$\pm$0.045&  0.666$\pm$0.269  & 5.547$\pm$0.405                     & 0.563$\pm$0.918                  & 21.874$\pm$1.162                 & 27.803$\pm$1.407                      \\ \cmidrule{2-8} 
\multicolumn{1}{c|}{}                              & $\alpha$=0.6  & 0.340$\pm$0.046
&  0.650$\pm$0.275  & 5.585$\pm$0.399                     & 0.486$\pm$0.960                  & 21.764$\pm$1.158                 & 27.785$\pm$1.446                      \\ \cmidrule{2-8} 
\multicolumn{1}{c|}{}                              & $\alpha$=0.8 & 0.343$\pm$0.045
&   0.661$\pm$0.268   & 5.582$\pm$0.436                     & 0.491$\pm$0.943                  & 21.821$\pm$1.169                 & 27.709$\pm$1.448                      \\ \midrule
\multicolumn{2}{c|}{Forward KL Divergence} & 0.344$\pm$0.046  &  
 0.664$\pm$0.266   &5.589$\pm$0.416                     & 0.517$\pm$0.942                  & 21.852$\pm$1.138                 & 27.854$\pm$1.446                      \\ \midrule
\multicolumn{2}{c|}{Jensen-Shannon Divergence} & 0.342$\pm$0.045 &  0.661$\pm$0.268  &
  \textbf{5.649$\pm$0.409}            & \textbf{0.573$\pm$0.940}         & \textbf{21.904$\pm$1.158}        & \textbf{27.880$\pm$1.436}             \\ \bottomrule
\end{tabular}
\caption{Evaluations of the alignment performance with \textbf{Gen-AI Benchmark} experiments, where the CLIP score and VQAScore evaluates image-text alignment performance, and the remaining four metrics evaluate human value alignment performance.}
\label{evaluations of alignment performance_Gen-AI}
\end{table*}

\begin{table*}[ht]\scriptsize
\centering
\begin{tabular}{cc|cccc}
\toprule
\multicolumn{2}{c|}{Model}                               & \multicolumn{1}{c|}{Image-Image CLIP score $\downarrow$} & \multicolumn{1}{c|}{Entropy 1D $\uparrow$} & \multicolumn{1}{c|}{Entropy 2D $\uparrow$} & \multicolumn{1}{c}{LPIPS $\uparrow$} \\ \midrule \midrule
\multicolumn{2}{c|}{Original Model}                      & 0.8358 $\pm$ 0.0916                         & 3.8889 $\pm$ 0.1875                 & 7.6543 $\pm$ 0.4906                 & 0.3031 $\pm$ 0.0388            \\ \midrule \midrule
\multicolumn{2}{c|}{Reverse KL Divergence}               & 0.8668 $\pm$ 0.0843                         & 3.9865 $\pm$ 0.1107                 & 7.8165 $\pm$ 0.3640                 & 0.3020 $\pm$ 0.0355            \\ \midrule
\multicolumn{1}{c|}{\multirow{5.20}{*}{$\alpha$-Divergence}} & $\alpha=$0.2 & 0.8683 $\pm$ 0.0834                         & 3.9724 $\pm$ 0.1298                 & 7.8136 $\pm$ 0.4026                 & 0.3121 $\pm$ 0.0371            \\ \cmidrule{2-6} 
\multicolumn{1}{c|}{}                              & $\alpha=$0.4 & 0.8646 $\pm$ 0.0857                         & \textbf{4.0095 $\pm$ 0.1018}        & 7.8478 $\pm$ 0.3396                 & 0.3058 $\pm$ 0.0336            \\ \cmidrule{2-6} 
\multicolumn{1}{c|}{}                              & $\alpha=$0.6 &  \textbf{0.8608 $\pm$ 0.0864}                & 3.9634 $\pm$ 0.1252                 & 7.8141 $\pm$ 0.3904                 & \textbf{0.3185 $\pm$ 0.0371 }   \\ \cmidrule{2-6} 
\multicolumn{1}{c|}{}                              & $\alpha=$0.8 & 0.8653 $\pm$ 0.0870                         & 3.9733 $\pm$ 0.1398                 & 7.7432 $\pm$ 0.4297                 & 0.3088 $\pm$ 0.0380            \\ \midrule
\multicolumn{2}{c|}{Forward KL Divergence}               & 0.8682 $\pm$ 0.0847                         & 3.9804 $\pm$ 0.1143                 & 7.7786 $\pm$ 0.3660                 & 0.3072 $\pm$ 0.0350            \\ \midrule
\multicolumn{2}{c|}{Jensen-Shannon Divergence}           & 0.8685 $\pm$ 0.0818                         & 3.9886 $\pm$ 0.1145                 & \textbf{7.8590 $\pm$ 0.3808}        & 0.3103 $\pm$ 0.0359            \\ \bottomrule
\end{tabular}
\\
\begin{tabular}{cc|cccc}
\toprule
\multicolumn{2}{c|}{Model}                               & \multicolumn{1}{c|}{\;\quad\quad RMSE $\uparrow$\quad\quad\;} & \multicolumn{1}{c|}{\quad\quad\quad PSNR $\downarrow$\quad\quad\quad} & \multicolumn{1}{c|}{\quad\quad\quad SSIM $\downarrow$\quad\quad\quad} & \multicolumn{1}{c}{\quad\quad FSIM $\downarrow$\quad\quad} \\ \midrule \midrule
\multicolumn{2}{c|}{Original Model}                      & 0.0133 $\pm$ 0.0026           & 37.686 $\pm$ 1.797            & 0.8838 $\pm$ 0.0323           & 0.3795 $\pm$ 0.0213           \\ \midrule \midrule
\multicolumn{2}{c|}{Reverse KL Divergence}               & 0.0153 $\pm$ 0.0025           & 36.415 $\pm$ 1.501            & 0.8550 $\pm$ 0.0326           & 0.3803 $\pm$ 0.0188           \\ \midrule
\multicolumn{1}{c|}{\multirow{4}{*}{$\alpha$-Divergence}} & $\alpha=$0.2 & 0.0161 $\pm$ 0.0025           & 35.943 $\pm$ 1.453            & 0.8445 $\pm$ 0.0339           & \textbf{0.3755 $\pm$ 0.0208}  \\ \cmidrule{2-6} 
\multicolumn{1}{c|}{}                              & $\alpha=$0.4 & 0.0155 $\pm$ 0.0023           & 36.288 $\pm$ 1.363            & 0.8540 $\pm$ 0.0308           & 0.3806 $\pm$ 0.0180           \\ \cmidrule{2-6} 
\multicolumn{1}{c|}{}                              & $\alpha=$0.6 & \textbf{0.0165 $\pm$ 0.0025}  & \textbf{35.757 $\pm$ 1.384}   & \textbf{0.8394 $\pm$ 0.0330}  & 0.3771 $\pm$ 0.0209        \\ \cmidrule{2-6} 
\multicolumn{1}{c|}{}                              & $\alpha=$0.8 & 0.0153 $\pm$ 0.0026           & 36.417 $\pm$ 1.632            & 0.8553 $\pm$ 0.0349           & 0.3783 $\pm$ 0.0210           \\ \midrule
\multicolumn{2}{c|}{Forward KL Divergence}               & 0.0157 $\pm$ 0.0026           & 36.170 $\pm$ 1.410            & 0.8501 $\pm$ 0.0315           & 0.3762 $\pm$ 0.0189           \\ \midrule
\multicolumn{2}{c|}{Jensen-Shannon Divergence}           & 0.0158 $\pm$ 0.0026           & 36.092 $\pm$ 1.372            & 0.8473 $\pm$ 0.0313          & 0.3793 $\pm$ 0.0201           \\ \bottomrule
\end{tabular}
\caption{Evaluations of the generation diversity with \textbf{Gen-AI Benchmark}. The metrics originally utilized for evaluating image similarity exhibit an opposite property when evaluating generation diversity.}
\label{evaluations of diversity performance_Gen-AI}
\end{table*}

\setlength{\tabcolsep}{1.5mm}
\begin{table*}[h]\scriptsize
\centering
\begin{tabular}{cc|cc|cccc}
\toprule
\multicolumn{2}{c|}{Model}& \multicolumn{1}{c|}{CLIPScore $\uparrow$}& \multicolumn{1}{c|}{VQAScore $\uparrow$} & \multicolumn{1}{c|}{Aesthetics Score $\uparrow$} & \multicolumn{1}{c|}{ImageReward $\uparrow$} & \multicolumn{1}{c|}{Pickscore $\uparrow$} &  {HPS-V2   $\uparrow$}                           \\ \midrule \midrule 
\multicolumn{2}{c|}{Original Model}  & 0.343$\pm$0.054 & 0.658$\pm$0.251 & 5.575$\pm$0.556                     & 0.231$\pm$1.047                   & 21.059$\pm$1.216                 & \multicolumn{1}{c}{27.082$\pm$1.541} \\ \midrule \midrule
\multicolumn{2}{c|}{Reverse KL Divergence}& \textbf{0.353$\pm$0.053} &  \textbf{0.710$\pm$0.234} & 5.698$\pm$0.534                     & 0.661$\pm$0.940                  & 21.682$\pm$1.194                 & 27.907$\pm$1.519                      \\ \midrule
\multicolumn{1}{c|}{\multirow{5.20}{*}{$\alpha$-Divergence}} & $\alpha$=0.2 & 0.352$\pm$0.054 &  0.705$\pm$0.236  & 5.712$\pm$0.527                     & 0.627$\pm$0.962                  & 21.581$\pm$1.200                 & 27.953$\pm$1.543                      \\ \cmidrule{2-8} 
\multicolumn{1}{c|}{}                              & $\alpha$=0.4 & 0.351$\pm$0.053&  0.700$\pm$0.239  & 5.659$\pm$0.519                     & 0.626$\pm$0.957                  & 21.611$\pm$1.205                 & 27.840$\pm$1.504                      \\ \cmidrule{2-8} 
\multicolumn{1}{c|}{}                              & $\alpha$=0.6  & 0.349$\pm$0.053
&  0.691$\pm$0.241  & 5.666$\pm$0.509                     & 0.569$\pm$0.971                  & 21.456$\pm$1.205                 & 27.831$\pm$1.513                      \\ \cmidrule{2-8} 
\multicolumn{1}{c|}{}                              & $\alpha$=0.8 & 0.351$\pm$0.054
&   0.697$\pm$0.239   & 5.701$\pm$0.537                     & 0.598$\pm$0.969                  & 21.555$\pm$1.195                 & 27.786$\pm$1.510                      \\ \midrule
\multicolumn{2}{c|}{Forward KL Divergence} & 0.353$\pm$0.054  &  
 0.706$\pm$0.236   &5.735$\pm$0.524                     & 0.626$\pm$0.952                  & 21.640$\pm$1.184                 & 27.941$\pm$1.510                      \\ \midrule
\multicolumn{2}{c|}{Jensen-Shannon Divergence} & 0.352$\pm$0.053 &  0.707$\pm$0.235  &
  \textbf{5.765$\pm$0.513}            & \textbf{0.672$\pm$0.942}         & \textbf{21.708$\pm$1.194}        & \textbf{27.954$\pm$1.502}             \\ \bottomrule
\end{tabular}
\caption{Evaluations of the alignment performance with larger-scale (parti-prompts and HPS-V2 training set, \textbf{total 4832 prompts}) experiments, where the CLIP score and VQAScore evaluates image-text alignment performance, and the remaining four metrics evaluate human value alignment performance.}
\label{evaluations of alignment performance_larger-scale}
\end{table*}

\begin{table*}[ht]\scriptsize
\centering
\begin{tabular}{cc|cccc}
\toprule
\multicolumn{2}{c|}{Model}                               & \multicolumn{1}{c|}{Image-Image CLIP score $\downarrow$} & \multicolumn{1}{c|}{Entropy 1D $\uparrow$} & \multicolumn{1}{c|}{Entropy 2D $\uparrow$} & \multicolumn{1}{c}{LPIPS $\uparrow$} \\ \midrule \midrule
\multicolumn{2}{c|}{Original Model}                      & 0.8096 $\pm$ 0.0982                         & 3.8279 $\pm$ 0.2675                 & 7.5427 $\pm$ 0.5896                 & 0.3002 $\pm$ 0.0404            \\ \midrule \midrule
\multicolumn{2}{c|}{Reverse KL Divergence}               & 0.8491 $\pm$ 0.0891                         & 3.9519 $\pm$ 0.1578                 & 7.7858 $\pm$ 0.3904                 & 0.2957 $\pm$ 0.0349            \\ \midrule
\multicolumn{1}{c|}{\multirow{5.20}{*}{$\alpha$-Divergence}} & $\alpha=$0.2 & 0.8420 $\pm$ 0.0887                         & 3.9231 $\pm$ 0.1743                 & \textbf{7.8985 $\pm$ 0.3214}                 & 0.3096 $\pm$ 0.0321            \\ \cmidrule{2-6} 
\multicolumn{1}{c|}{}                              & $\alpha=$0.4 & \textbf{0.8471 $\pm$ 0.0892}                         & \textbf{3.9689 $\pm$ 0.1564}        & 7.7860 $\pm$ 0.3721                 & 0.2995 $\pm$ 0.0348            \\ \cmidrule{2-6} 
\multicolumn{1}{c|}{}                              & $\alpha=$0.6 & \textbf{0.8472 $\pm$ 0.0881}                & 3.9216 $\pm$ 0.1932                 & 7.7592 $\pm$ 0.4400                 & \textbf{0.3132 $\pm$ 0.0140 }   \\ \cmidrule{2-6} 
\multicolumn{1}{c|}{}                              & $\alpha=$0.8 & 0.8423 $\pm$ 0.0795                         & 3.9448 $\pm$ 0.1851                 & 7.7059 $\pm$ 0.4409                 & 0.3030 $\pm$ 0.0363            \\ \midrule
\multicolumn{2}{c|}{Forward KL Divergence}               & 0.8505 $\pm$ 0.0876                         & 3.9409 $\pm$ 0.1639                 & 7.7315 $\pm$ 0.3868                 & 0.3004 $\pm$ 0.0346            \\ \midrule
\multicolumn{2}{c|}{Jensen-Shannon Divergence}           & 0.8503 $\pm$ 0.0885                         & 3.9532 $\pm$ 0.1569                 & 7.8239 $\pm$ 0.3928        & 0.3036 $\pm$ 0.0358            \\ \bottomrule
\end{tabular}
\\
\begin{tabular}{cc|cccc}
\toprule
\multicolumn{2}{c|}{Model}                               & \multicolumn{1}{c|}{\;\quad\quad RMSE $\uparrow$\quad\quad\;} & \multicolumn{1}{c|}{\quad\quad\quad PSNR $\downarrow$\quad\quad\quad} & \multicolumn{1}{c|}{\quad\quad\quad SSIM $\downarrow$\quad\quad\quad} & \multicolumn{1}{c}{\quad\quad FSIM $\downarrow$\quad\quad} \\ \midrule \midrule
\multicolumn{2}{c|}{Original Model}                      & 0.0133 $\pm$ 0.0028           & 37.660 $\pm$ 1.879            & 0.8819 $\pm$ 0.0380           & 0.3779 $\pm$ 0.0227           \\ \midrule \midrule
\multicolumn{2}{c|}{Reverse KL Divergence}               & 0.0154 $\pm$ 0.0027           & 36.360 $\pm$ 1.587            & 0.8521 $\pm$ 0.0374           & 0.3800 $\pm$ 0.0185           \\ \midrule
\multicolumn{1}{c|}{\multirow{4}{*}{$\alpha$-Divergence}} & $\alpha=$0.2 & \textbf{0.0176 $\pm$ 0.0025}           & \textbf{35.192 $\pm$ 1.280}            & \textbf{0.8354 $\pm$ 0.0350}           & \textbf{0.3757 $\pm$ 0.0182}  \\ \cmidrule{2-6} 
\multicolumn{1}{c|}{}                              & $\alpha=$0.4 & 0.0155 $\pm$ 0.0026           & 36.317 $\pm$ 1.489            & 0.8535 $\pm$ 0.0357           & 0.3806 $\pm$ 0.0184           \\ \cmidrule{2-6} 
\multicolumn{1}{c|}{}                              & $\alpha=$0.6 & 0.0166 $\pm$ 0.0027  & 35.708 $\pm$ 1.488   & 0.8371 $\pm$ 0.0348  & 0.3765 $\pm$ 0.0212        \\ \cmidrule{2-6} 
\multicolumn{1}{c|}{}                              & $\alpha=$0.8 & 0.0155 $\pm$ 0.0028           & 36.327 $\pm$ 1.646            & 0.8528 $\pm$ 0.0383           & 0.3789 $\pm$ 0.0210           \\ \midrule
\multicolumn{2}{c|}{Forward KL Divergence}               & 0.0157 $\pm$ 0.0026           & 36.133 $\pm$ 1.536            & 0.8475 $\pm$ 0.0366           & 0.3763 $\pm$ 0.0194           \\ \midrule
\multicolumn{2}{c|}{Jensen-Shannon Divergence}           & 0.0158 $\pm$ 0.0026           & 36.054 $\pm$ 1.511            & 0.8450 $\pm$ 0.0363          & 0.3798 $\pm$ 0.0202           \\ \bottomrule
\end{tabular}
\caption{Evaluations of the generation diversity with larger-scale (parti-prompts and HPS-V2 training set, \textbf{total 4832 prompts}) experiments. The metrics originally utilized for evaluating image similarity exhibit an opposite property when evaluating generation diversity.}
\label{generation diversity evaluation_larger-scale}
\end{table*}
\clearpage
\section{Qualitative Comparison of Alignment with Diverse Divergence}
\begin{figure}[ht]
\centering
\includegraphics[width=1.0\columnwidth]{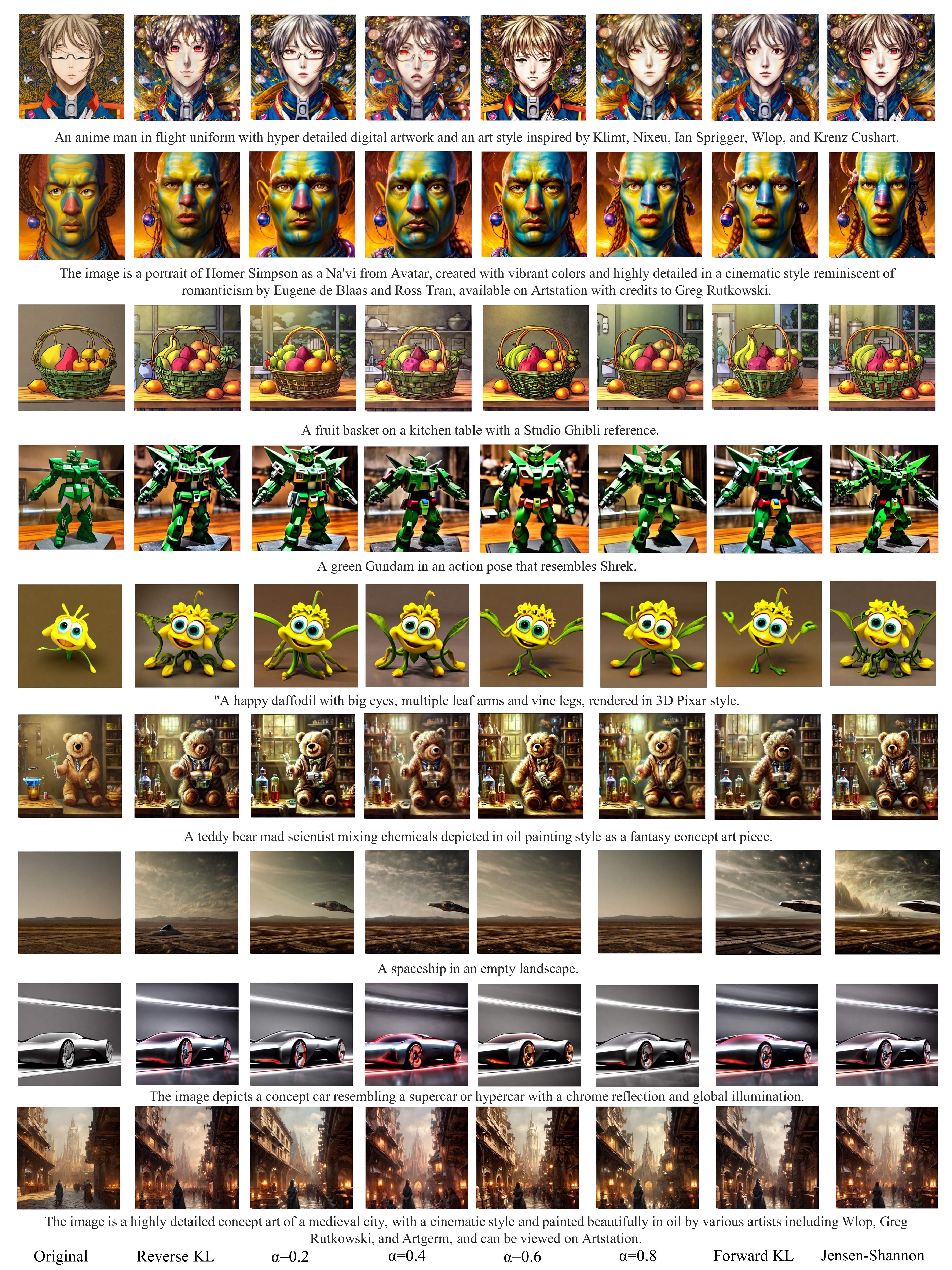}
\label{examples_2}
\end{figure}

\begin{figure}[ht]
\centering
\includegraphics[width=1.0\columnwidth]{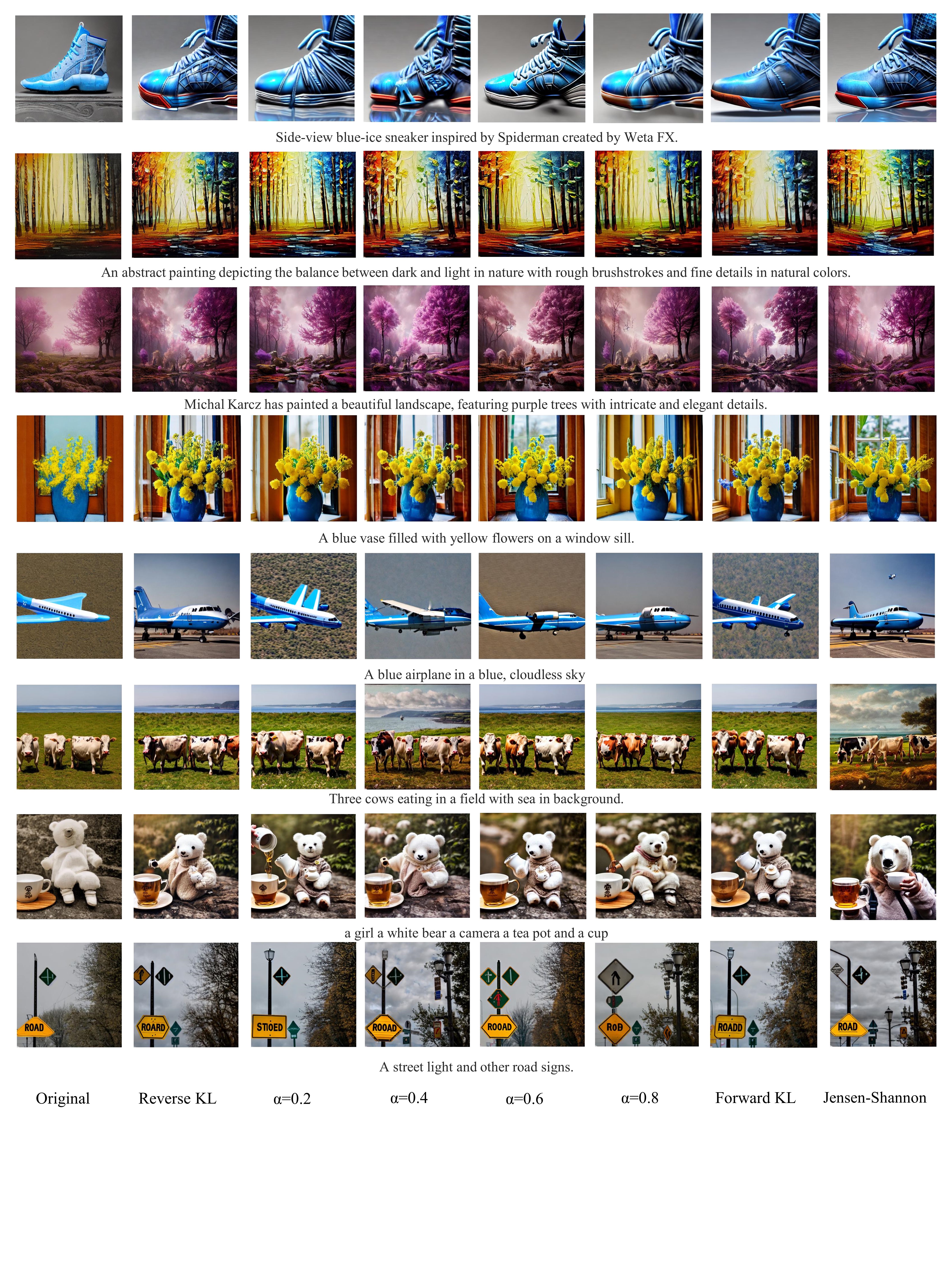}
\label{examples_3}
\end{figure}
\clearpage

\begin{figure}[H]
\centering
\includegraphics[width=1.0\columnwidth]{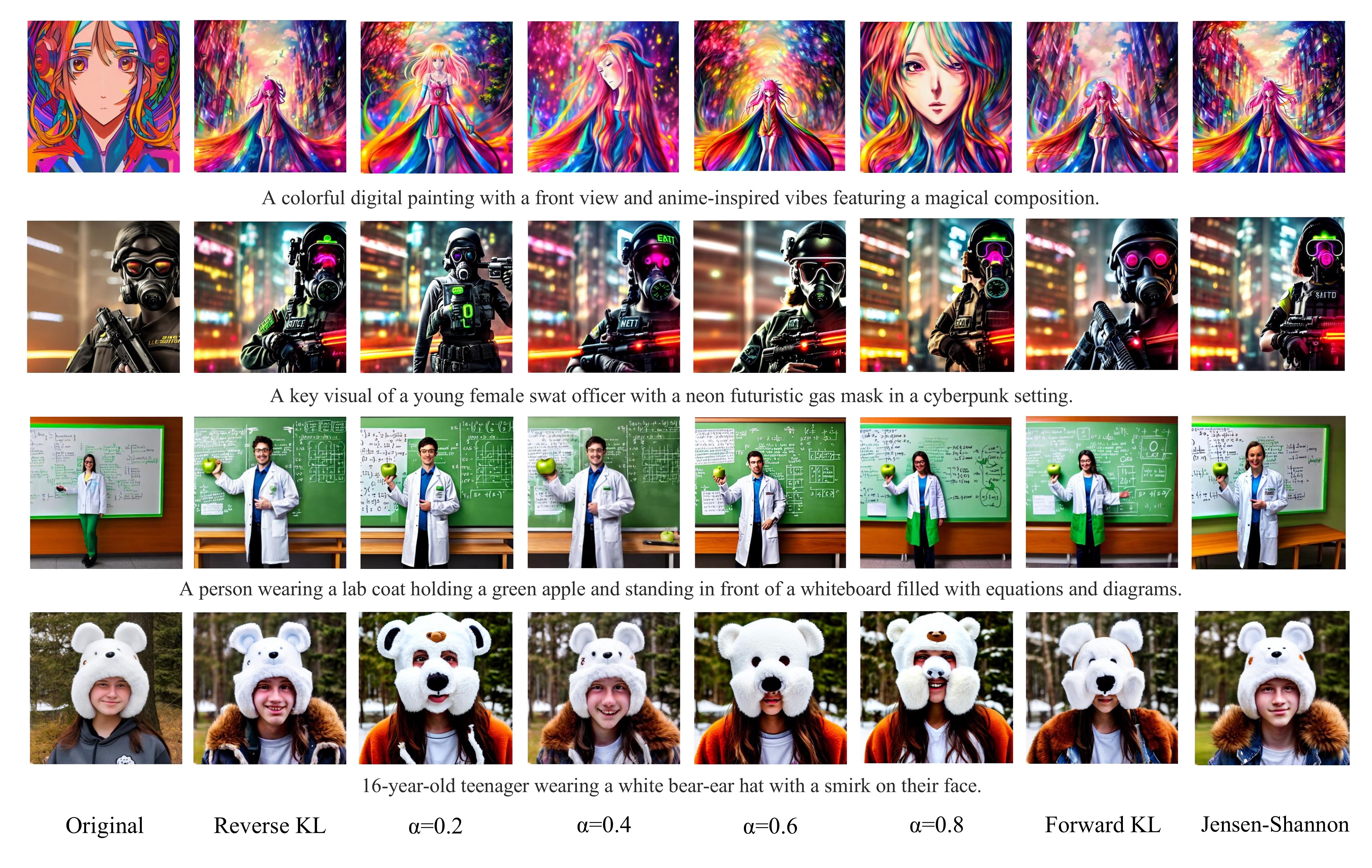}
\label{examples_1}
\end{figure}

\end{document}